\documentclass[sigconf]{acmart}
\usepackage{balance}
\usepackage{enumitem}
\usepackage{microtype}
\usepackage{array}
\usepackage{multirow}
\usepackage{longtable}
\usepackage{booktabs}
\usepackage{xcolor}
\usepackage{colortbl,booktabs}
\usepackage{color}
\usepackage{graphicx} 
\usepackage{subfigure}
\usepackage{balance}
\usepackage{bbm}
\usepackage{tikz}

\newcommand\BibTeX{B\textsc{ib}\TeX}
\AtBeginDocument{%
  \providecommand\BibTeX{{%
    Bib\TeX}}}

\AtBeginDocument{%
  \providecommand\BibTeX{{%
    \normalfont B\kern-0.5em{\scshape i\kern-0.25em b}\kern-0.8em\TeX}}}

\setcopyright{acmcopyright}
\copyrightyear{2018}
\acmYear{2018}
\acmDOI{XXXXXXX.XXXXXXX}

\acmConference[Conference XX]{Make sure to enter the correct
  conference title from your rights confirmation emai}{2024}{}
\acmPrice{15.00}
\acmISBN{978-1-4503-XXXX-X/18/06}

\acmSubmissionID{1773}

\usepackage{soul}
\begin{document}
\title{\textsc{MM-BigBench}: Evaluating Multimodal Models on Multimodal Content Comprehension Tasks}
\author{Xiaocui Yang}
\authornote{Xiaocui and Wenfang have equal contribution.}
\affiliation{%
  \institution{Northeastern University, China}
  \country{}
}
\email{yangxiaocui@stumail.neu.edu.cn}

\author{Wenfang Wu}
\affiliation{%
  \institution{Northeastern University, China}
  \country{}
}
\email{wenfang@stumail.neu.edu.cn}

\author{Shi Feng}
\affiliation{%
  \institution{Northeastern University, China}
  \country{}
}
\email{fengshi@cse.neu.edu.cn}

\author{Ming Wang}
\affiliation{%
  \institution{Northeastern University, China}
  \country{}
}
\email{wangming@stumail.neu.edu.cn}

\author{Daling Wang}
\affiliation{%
  \institution{Northeastern University, China}
    \country{}
}
\email{wangdaling@cse.neu.edu.cn}

\author{Yang Li}
\affiliation{%
  \institution{Northeastern University, China}
  \country{}
}
\email{liyang@stumail.neu.edu.cn}

\author{Qi Sun}
\affiliation{%
  \institution{Nanjing University of Science and Technology, China}
  \country{}
}
\email{319106003718@njust.edu.cn}

\author{Yifei Zhang}
\affiliation{%
  \institution{Northeastern University, China}
    \country{}
}
\email{zhangyifei@cse.neu.edu.cn}

\author{Xiaoming Fu}
\affiliation{%
  \institution{University of Göttingen, Germany}
    \country{}
}
\email{fu@cs.uni-goettingen.de}

\author{Soujanya Poria}
\affiliation{%
  \institution{Singapore University of Technology and Design, Singapore}
  \country{}
}
\email{sporia@sutd.edu.sg}

\renewcommand{\shortauthors}{Xiaocui Yang and Wenfang Wu, et al.}
\renewcommand\footnotetextcopyrightpermission[1]{}
\settopmatter{printacmref=false}

\begin{abstract}
The popularity of multimodal large language models (MLLMs) has triggered a recent surge in research efforts dedicated to evaluating these models.
Nevertheless, existing evaluation studies of MLLMs primarily focus on the comprehension and reasoning of unimodal (vision) content, neglecting performance evaluations in the domain of multimodal (vision-language) content understanding.
Beyond multimodal reasoning, tasks related to multimodal content comprehension necessitate a profound understanding of multimodal contexts, achieved through the multimodal interaction to obtain a final answer.
In this paper, we introduce a comprehensive assessment framework called \textbf{\textsc{MM-BigBench}}, which incorporates a diverse range of metrics to offer an extensive evaluation of the performance of various models and instructions across a wide spectrum of diverse \textbf{multimodal content comprehension tasks}.
Consequently, our work complements research on the performance of MLLMs in multimodal comprehension tasks, achieving a more comprehensive and holistic evaluation of MLLMs.
To begin, we employ the `\textbf{Best Performance}' metric to ascertain each model's performance upper bound on different datasets. Subsequently, the `\textbf{Mean Relative Gain}' metric offers an assessment of the overall performance of various models and instructions, while the `\textbf{Stability}' metric measures their sensitivity.
Furthermore, previous research centers on evaluating models independently or solely assessing instructions, neglecting the adaptability between models and instructions. We propose the `\textbf{Adaptability}' metric to quantify the adaptability between models and instructions.
Our paper evaluates a total of 20 language models (14 MLLMs) on 14 multimodal datasets spanning 6 tasks, with 10 instructions for each task, and derives novel insights. Our code will be released at \url{https://github.com/declare-lab/MM-BigBench}.
\end{abstract}





\maketitle
\begin{tikzpicture}[remember picture,overlay,shift={(current page.north west)}]
\node[anchor=north west,xshift=1.1cm,yshift=-2.5cm]{\scalebox{-1}[1]{\includegraphics[width=1cm]{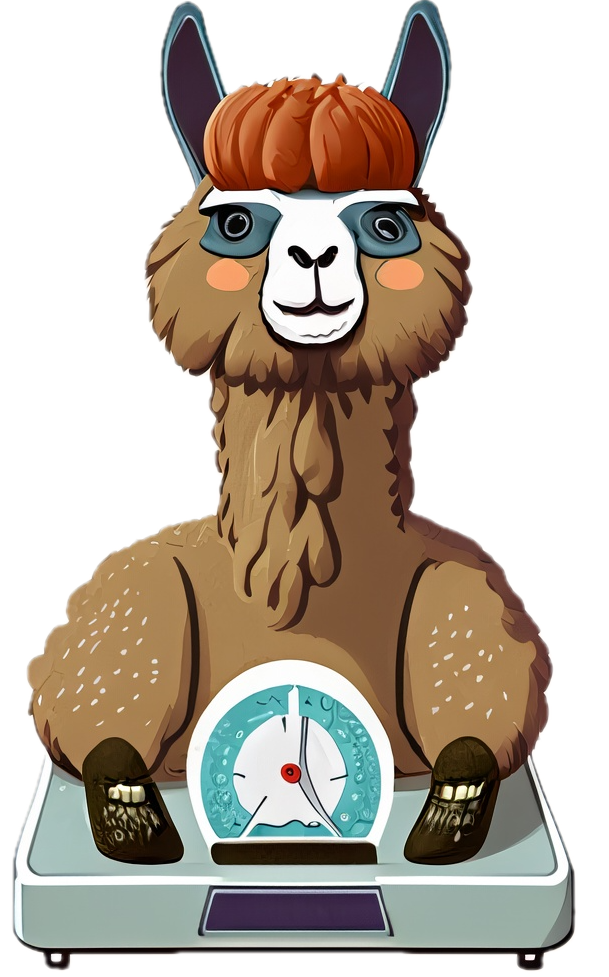}}};
\end{tikzpicture}
\section{Introduction}
\begin{figure*}[t] 
  \centering 
  \includegraphics[width = 0.9\textwidth]{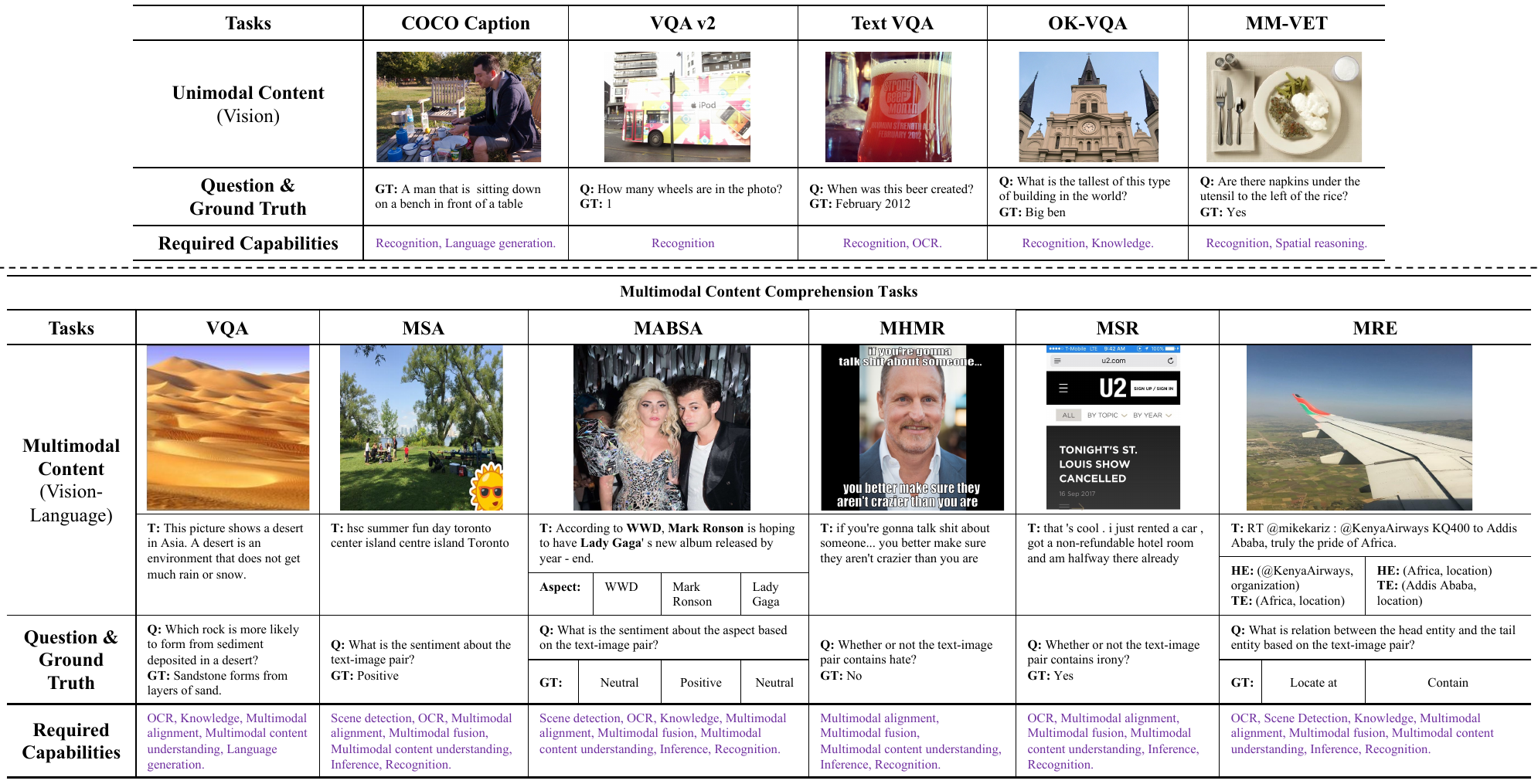}
  \caption{Required Capabilities for diverse benchmarks, e.g., COCO \cite{DBLP:conf/eccv/LinMBHPRDZ14}, VQA v2 \cite{DBLP:conf/eccv/LinMBHPRDZ14}, Text VQA \cite{DBLP:conf/cvpr/SinghNSJCBPR19}, OK-VQA \cite{DBLP:conf/cvpr/MarinoRFM19}, MM-VET \cite{DBLP:journals/corr/abs-2308-02490}, and our multimodal content comprehension tasks. Multimodal content comprehension tasks (below the dotted line) require not only interactions between the multimodal content to explore traditional vision-language multimodal capabilities, such as knowledge reasoning, spatial reasoning, OCR recognition, and more, but also a deep understanding of multimodal content. 
  `T' represents the text content, `Q' denotes the question used to prompt models for answers, and `GT' stands for the ground truth label. `HE' corresponds to the head entity, while `TE' corresponds to the tail entity.
   }
  \label{Figure_Multimodal Content Comprehension Tasks} 
\end{figure*}
Multimodal Large Language Models (MLLMs), harnessing the formidable capabilities of Large Language Models (LLMs), demonstrate outstanding performance across a spectrum of multimodal tasks~\cite{DBLP:conf/nips/AlayracDLMBHLMM22, DBLP:journals/corr/abs-2304-15010, DBLP:journals/corr/abs-2301-12597, DBLP:journals/corr/abs-2305-06500}. 
The emergence of recent research\footnote{The detailed introduction to related research on the evaluation of LMs can be found in the Related Work section of the Appendix \ref{Appendix_related_works}.}, including but not limited to MME \cite{DBLP:journals/corr/abs-2306-13394}, SEED-Bench \cite{DBLP:journals/corr/abs-2307-16125}, LVLM-eHub \cite{DBLP:journals/corr/abs-2306-09265}, and MM-Vet \cite{DBLP:journals/corr/abs-2308-02490}, has predominantly focused on appraising the required traditional vision-language multimodal capabilities of MLLMs in the tasks primarily driven by visual content (vision+text question) like Visual Question Answering (VQA) and Video Question Answering (VideoQA).
These capabilities encompass recognition, OCR, spatial, knowledge, math reasoning, and etc., as depicted in the section above the dotted line in Figure \ref{Figure_Multimodal Content Comprehension Tasks}.
However, there is limited understanding about the performance of MLLMs in multimodal content comprehension tasks (vison-language+text question), such as Multimodal Sentiment Analysis (MSA) \cite{DBLP:conf/mmm/NiuZPE16, DBLP:journals/tmm/YangFW021, DBLP:journals/corr/ZadehZPM16, DBLP:conf/acl/MorencyCPLZ18}, Multimodal Aspect-Based Sentiment Analysis (MABSA) \cite{DBLP:conf/aaai/0001FLH18, DBLP:conf/acl/JiZCLN18, DBLP:journals/ijon/ZhouZHHH21}, Multimodal Hateful Memes Recognition (MHMR) \cite{mathias2021findings}, Multimodal Sarcasm Recognition (MSR) \cite{DBLP:conf/acl/CaiCW19}, Multimodal Relation Extraction (MRE) \cite{mathias2021findings}, and the VQA with text context \cite{DBLP:conf/nips/LuMX0CZTCK22}.
On the other hand, in Natural Language Processing (NLP), most research \cite{DBLP:journals/corr/abs-2305-15005, DBLP:journals/corr/abs-2304-04339, DBLP:journals/corr/abs-2308-04945} primarily focuses on evaluating pure LLMs like ChatGPT \cite{chatgpt}, Flan-T5 \cite{DBLP:journals/corr/abs-2210-11416}, and others, specifically for text classification tasks such as text sentiment analysis, and relation classification. This leaves the performance of various MLLMs in multimodal content comprehension tasks that rely on text and image modalities largely unexplored.

To address the aforementioned gap, we conduct the comprehensive evaluation involving 20 publicly available models (as listed in Table \ref{Table_different_models}), including 14 MLLMs, across a diverse set of 14 datasets covering 6 distinct tasks (as shown in Table \ref{Table_dataset_statistic}). Our primary focus is to assess the performance of various MLLMs in the context of tasks involving the comprehension of multimodal content, specifically text-image pairs. We also aim to establish benchmarks across a range of MLLMs for diverse multimodal content comprehension tasks. These tasks not only require conventional vision-language multimodal  capabilities in the models but also demand a deep understanding of multimodal content for classification (sentiment analysis, hate speech, sarcasm, etc.) or reasoning (visual question answering), as displayed in the section below the dotted line in Figure \ref{Figure_Multimodal Content Comprehension Tasks}.  The comprehension of multimodal content frequently necessitates various cognitive processes, including but not limited to modality alignment and multimodal fusion, as highlighted in studies such as \cite{DBLP:conf/emnlp/HanCP21} and \cite{DBLP:conf/mm/HazarikaZP20}, in addition to semantic understanding. It's worth noting that precisely defining the boundaries of such multimodal content comprehension can be challenging, given that the content may encompass diverse elements, such as language-visual alignment through OCR, and spatial reasoning, among others. Furthermore, when it comes to visual question answering in the context of multimodal content, particularly with prompts or instructions, it often triggers these reasoning abilities, as vividly demonstrated in datasets like ScienceQA \cite{DBLP:conf/nips/LuMX0CZTCK22}.

We introduce the comprehensive assessment framework called \textbf{\textsc{MM-BigBench}}, incorporating a diverse set of metrics to conduct a thorough evaluation of various models and instructions in the context of multimodal content comprehension tasks. MM-BigBench serves as a complement to existing evaluation studies of MLLMs, offering a more comprehensive and holistic assessment when combined with prior related work.
Specifically, we propose the `\textbf{Best Performance}' metric to gain insights into how each model attains its highest performance on a specific instruction for each dataset. This metric represents the upper limit and benchmark of each model's performance on each dataset.
To evaluate the overall performance of each model across all instructions on a specific dataset, we introduce the `\textbf{Model Mean Relative Gain}' metric. Likewise, to assess the overall performance of each instruction across all models, the `\textbf{Instruction Mean Relative Gain}' metric is introduced.
Stability is also a crucial indicator of model and instruction performance. A model that consistently performs well across all instructions is considered more stable, and an instruction that exhibits strong performance across all models holds a significant advantage. We introduce the `\textbf{Model Stability}' and `\textbf{Instruction Stability}' metrics to assess this aspect.
Furthermore, prior studies either focus solely on assessing different models \cite{DBLP:journals/corr/abs-2306-13394, DBLP:journals/corr/abs-2307-16125, DBLP:journals/corr/abs-2306-09265, DBLP:journals/corr/abs-2308-02490, DBLP:journals/corr/abs-2306-04757} or purely evaluate the performance of instructions \cite{DBLP:journals/corr/abs-2307-00259}, overlooking the issue of adaptability between models and instructions. 
To address this gap, we propose the `\textbf{Adaptability}' metric to quantify the adaptability between different models and various instructions. This metric measures the proportion of times each instruction achieves top-K performance on a specific model across all datasets. With this metric, we can discern which specific instructions yield superior performance for each model. 
We conduct extensive experiments, and the results demonstrate the following: (1) In multimodal content comprehension tasks, LMs with Flan-T5-XXL \cite{DBLP:journals/corr/abs-2210-11416} as the backbone, which is based on the Encoder-Decoder architecture
outperform the LLaMA series models \cite{DBLP:journals/corr/abs-2302-13971, DBLP:journals/corr/abs-2307-09288} with the Decoder-only architecture. (2) Instructions in a `Question-Answer' format perform better, and the addition of options further improves the model's performance on certain tasks.
(3) The performance of MLLMs, trained using instruction tuning, exhibits greater stability across various tasks when compared to models that do not undergo instruction tuning.
Our main contributions are summarized as follows:
\begin{itemize} [leftmargin=*]
\item{We conduct evaluations involving 20 models using 10 different instructions on 14 datasets, covering 6 multimodal content comprehension tasks. 
To our knowledge, this is the first work of the comprehensive assessment of a wide range of MLLMs using various manually designed instructions for diverse multimodal content comprehension tasks. Our work complements previous evaluation research on MLLMs and provides a more comprehensive and holistic assessment of MLLMs.
}
\item{We introduce the comprehensive assessment framework, \textbf{\textsc{MM-BigBench}}, with a range of diverse metrics to provide a thorough evaluation of different models and instructions, including the Best Performance metric, the Mean Relative Gain metric, the Stability metric, and the Adaptability metric.}
\item{
We conduct extensive experiments and establish benchmarks for LLMs and MLLMs on various multimodal content comprehension tasks, resulting in innovative conclusions.
}
\end{itemize}
\section{Multimodal Tasks and Datasets}
\subsection{Definition of Multimodal Content Comprehension Tasks}
Prior research in multimodal evaluation, as exemplified by references \cite{DBLP:journals/corr/abs-2306-13394, DBLP:journals/corr/abs-2307-16125, DBLP:journals/corr/abs-2306-09265, DBLP:journals/corr/abs-2308-02490}, confines MLLM evaluation to a limited understanding of their true multimodal capabilities. It primarily accentuates the interaction between textual instructions/questions and images (the visual modality), testing conventional vision-language multimodal capabilities, including knowledge reasoning, spatial reasoning, OCR, recognition, and more.
It's essential to note that these tasks indeed demand substantial multimodal reasoning skills, as they entail the deduction of information from text, aligning it with the image, and subsequently engaging in reasoning processes that inherently necessitate multimodal thinking. A notable limitation of such evaluations is the omission of assessing MLLMs in their ability to comprehend multimodal content.

\begin{figure*}[t] 
  \centering 
  \includegraphics[width=0.9\textwidth]{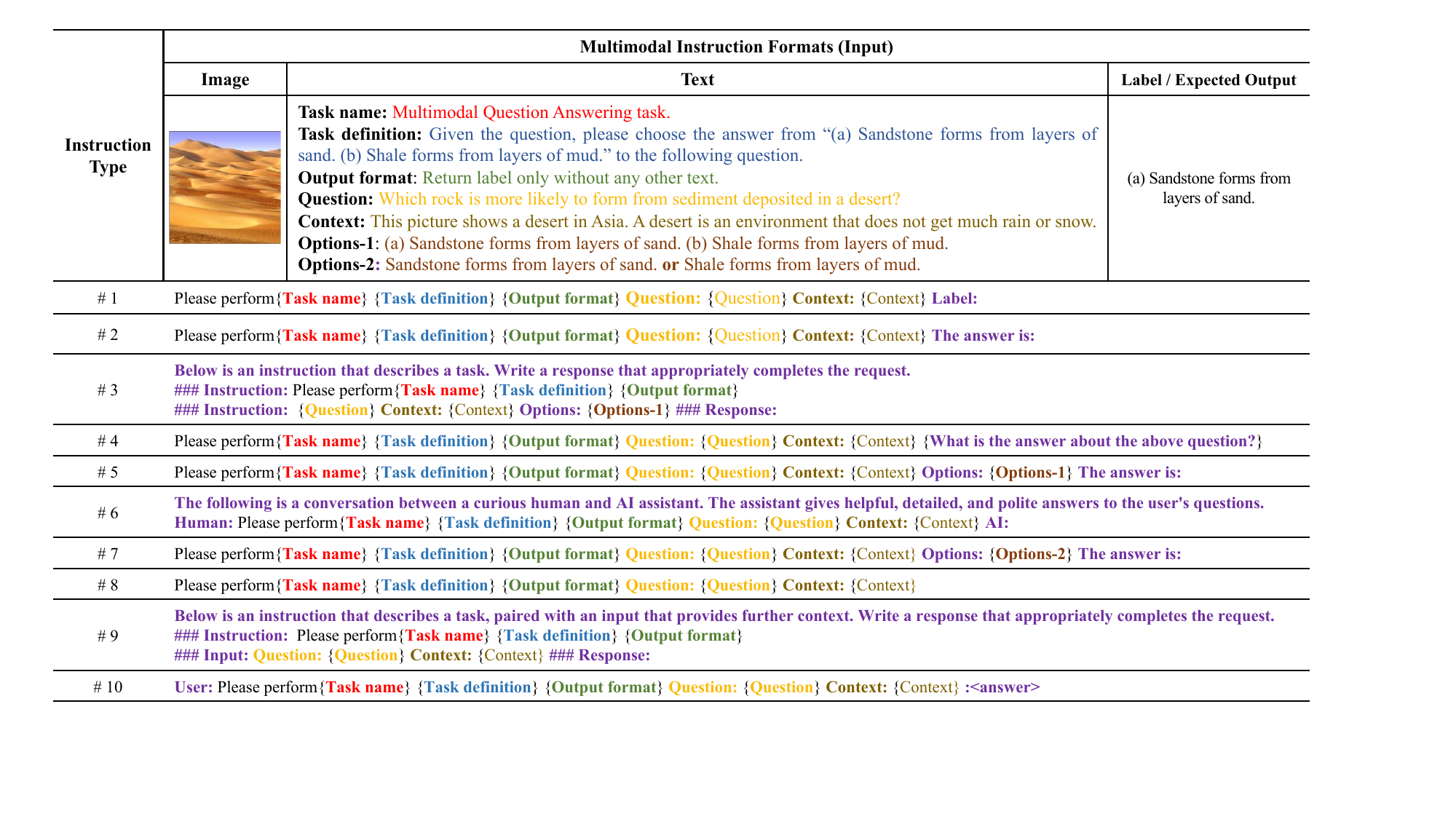}
  \vspace{-1em}
  \caption{
The column labeled "Multimodal Instruction Formats" displays the formats of different instructions designed specifically for ScienceQA. 
We construct instructions based on these formats, encompassing mandatory components, such as \textbf{\textcolor{red}{Task name}}, \textbf{\textcolor[RGB]{70,130,180}{Task definition}}, and \textbf{\textcolor[RGB]{34,139,34}{Output format}}, \textbf{\textcolor[RGB]{255,215,0}{Question}},
  as well as optional components, for instance, \textbf{\textcolor[RGB]{128,128,0}{Context}} and \textbf{\textcolor[RGB]{205,133,63}{Options}}. 
  Furthermore, each format incorporates \textbf{\textcolor[RGB]{153,50,204}{Specific instruction trigger words}} customized for the respective instruction.
   }
  \label{Figure_1_ScienceQA_different_instructions} 
\end{figure*}

Conversely, our research centers on the exploration of tasks related to the comprehension of multimodal content, such as multimodal content classification, which includes tasks like sentiment analysis and hate speech classification, as well as visual question-answering tasks. These tasks necessitate not only interactions between textual instructions and images for encompassing vision language multimodal capabilities as stated above but also a deep understanding of multimodal content, as exemplified in Figure \ref{Figure_Multimodal Content Comprehension Tasks}. An illustrative example is provided in Figure \ref{Figure_1_ScienceQA_different_instructions} and Figure \ref{Figure_1_MSA_Multimodal_Instruction}, where both textual descriptions and images are indispensable for formulating a response to the provided instruction.
Our aim is to conduct a comprehensive evaluation of various pretrained MLLMs across a diverse range of multimodal content comprehension tasks, encompassing MSA, MABSA, MSR, MHMR, MRE, and VQA. While the first five tasks focus on multimodal fusion or aligning textual contents with visual images, the last task focuses on traditional visual language capability testing such as recognition, spatial reasoning, OCR, knowledge utilization, etc. Like any other classification task, these multimodal content classification tasks do embrace the semantic understanding challenge.  Since our work involves evaluating MLLMs, it inherently covers multimodal reasoning by leveraging interactions between textual instructions and images, similar to prior works such as \cite{DBLP:journals/corr/abs-2308-02490, DBLP:journals/corr/abs-2306-13394, DBLP:journals/corr/abs-2307-16125, DBLP:journals/corr/abs-2306-09265, DBLP:journals/corr/abs-2306-04757}.


\subsection{Multimodal Datasets}
We conduct comprehensive evaluation of various Language Models across a diverse range of multimodal content comprehension tasks, including \textbf{MSA}, \textbf{MABSA}, \textbf{MHMR}, \textbf{MSR}, \textbf{MRE}, and \textbf{VQA}. Detailed statistics for each task and the datasets can be found in Table \ref{Table_dataset_statistic}. The detailed introduction to various multimodal content comprehension tasks and related datasets can be found in Appendix \ref{Appendex_datasets}.
\begin{table}[!htp] \small
\begin{center}
\renewcommand{\arraystretch}{1} 
\caption{Evaluation datasets statistics for different tasks. ``Labels'' denotes the number of labels among each task. `-' means that ScienceQA does not have a fixed label space, and each question has a unique answer.
}
\resizebox{1\linewidth}{!}{
\begin{tabular}{
p{0.85cm}< \centering| p{2.3cm}< \centering| p{1.4cm}< \centering |
p{0.5cm}< \centering| p{0.62cm}< \centering| p{0.6cm}< \centering }
\toprule[1pt]
\textbf{Task} & \textbf{Dataset} & \textbf{Modality} & \textbf{Test} & \textbf{Labels} & \textbf{Metric}\\
\midrule[1pt]
\multirow{7}{*}{\textbf{MSA}}
& MVSA-Single \cite{DBLP:conf/mmm/NiuZPE16} & Text-Image &413 & 3 & Acc \\
& MVSA-Multiple \cite{DBLP:conf/mmm/NiuZPE16} & Text-Image & 1531 & 3 & Acc \\
& TumEmo \cite{DBLP:journals/tmm/YangFW021} & Text-Image & 9463  & 7 & Acc \\
& MOSI-2 \cite{DBLP:journals/corr/ZadehZPM16} & Video & 654 & 2 & Acc\\
& MOSI-7 \cite{DBLP:journals/corr/ZadehZPM16} & Video & 684 & 7 & Acc \\
& MOSEI-2 \cite{DBLP:conf/acl/MorencyCPLZ18} & Video & 2797 & 2 & Acc\\
& MOSEI-7 \cite{DBLP:conf/acl/MorencyCPLZ18} & Video & 3588 & 7 & Acc \\
\midrule[0.5pt]
\multirow{3}{*}{\textbf{MABSA}}
& Twitter-2015 \cite{DBLP:conf/aaai/0001FLH18} & Text-Image & 1037 & 3 & Acc \\
& Twitter-2017 \cite{DBLP:conf/acl/JiZCLN18} & Text-Image & 1234 & 3 & Acc \\
& MASAD \cite{DBLP:journals/ijon/ZhouZHHH21} & Text-Image & 4935 & 2 & Acc \\
\midrule[0.5pt]
\textbf{MHMR}  & Hate \cite{mathias2021findings} & Text-Image & 500 & 2 & Acc \\
\midrule[0.5pt]
\textbf{MSR} & Sarcasm  \cite{DBLP:conf/acl/CaiCW19} & Text-Image & 2409 & 2 & Acc \\
\midrule[0.5pt]
\textbf{MRE}  & MNRE \cite{mathias2021findings} & Text-Image & 640 & 19 & Acc \\
\midrule[0.5pt]
\textbf{VQA} & ScienceQA \cite{DBLP:conf/nips/LuMX0CZTCK22} & Text-Image & 2017 & -  & Acc \\
\bottomrule[1pt]
\end{tabular}
}
\label{Table_dataset_statistic}
\end{center}
\end{table}

\section{Comprehensive Evaluation}
\subsection{Multimodal Instructions}
Recent research show that in the field of Natural Language Processing (NLP), different instructions, even when they have the same semantics, can result in significant differences in performance of a model \cite{DBLP:conf/nips/PerezKC21, DBLP:conf/acl/LuBM0S22, DBLP:journals/corr/abs-2305-15005, DBLP:journals/csur/LiuYFJHN23, gan2023sensitivity} .
Following \cite{DBLP:journals/corr/abs-2305-15005}, we design multimodal instructions that include text context, image context, essential components such as the task name, task definition, and output format, as well as optional components like answer options. The text context and image context constitute the multimodal content of each instance in various multimodal content comprehension tasks.
The \textbf{\textcolor{red}{Task name}} component serves to identify and define the purpose of each multimodal task. 
The \textbf{\textcolor[RGB]{70,130,180}{Task definition}} component is constructed based on the task's definition and annotation guidelines, and it includes the label space as a set of options for the model to generate its responses. 
The \textbf{\textcolor[RGB]{34,139,34}{Output format}} component specifies the expected structure of the model's output, facilitating the decoding of the model's response into the desired format.
The terms \textbf{\textcolor[RGB]{255,215,0}{Question}} and \textbf{\textcolor[RGB]{128,128,0}{Context}} are crucial in ScienceQA, and the ``Context'' term can be regarded as text context for ScienceQA.
In addition to the QA task, the optional ``Question'' component is utilized to simulate a Question-Answering (QA) task, and there is no context component.
The presence of the \textbf{\textcolor[RGB]{205,133,63}{Options}} term is optional and is used to provide the model with multiple-choice questions, prompting it to respond accordingly.
We notice that various Multimodal Large Language Models (MLLMs) have preferences for different instructions. This preference might be influenced by the use of \textbf{\textcolor[RGB]{153,50,204}{Specific instruction trigger words}} during their training, like ``\textbf{\#\#\# Instruction:}'', ``\textbf{Question:}''. With this understanding, we design a total of 10 instructions for each dataset. These instructions are created to evaluate how different MLLMs perform in a zero-shot manner when presented with various instructions.
We develop a variety of multimodal instructions for the ScienceQA task, as illustrated in Figure \ref{Figure_1_ScienceQA_different_instructions}.
Similar instructions are created for other tasks, maintaining a consistent structure that involves both text and image modalities. We also list the designed instructions for MSA, as shown in Figure \ref{Figure_1_MSA_Multimodal_Instruction} in Appendix \ref{Appendix_MSA_Multimodal_Instructions}. 
For detailed text instructions for various tasks, please refer to Figure \ref{Figure_2_text_instruction_for_different_tasks} in Appendix \ref{Appendix_different_instructions_for_multiple_taks}, which provides comprehensive details.

\subsection{Language Models}

We assess a diverse set of Language Models (LMs), including \textbf{6} pure LLMs, including \textbf{ChatGPT} \cite{chatgpt}, \textbf{Flan-T5} \cite{DBLP:journals/corr/abs-2210-11416}, \textbf{LLaMA-1-7B}, \textbf{LLaMA-1-13B} \cite{DBLP:journals/corr/abs-2302-13971} and \textbf{LLaMA-2-7B}, \textbf{LLaMA-2-13B} \cite{DBLP:journals/corr/abs-2307-09288}; \textbf{14} MLLMs, such as \textbf{OpenFlamingo} \cite{DBLP:conf/nips/AlayracDLMBHLMM22, anas_awadalla_2023_7733589}, \textbf{Fromage} \cite{koh2023grounding}, \textbf{LLaVA-7B}, \textbf{LLaVA-13B} \cite{DBLP:journals/corr/abs-2304-08485}, \textbf{MiniGPT-4} \cite{DBLP:journals/corr/abs-2304-10592}, \textbf{mPLUG-Owl} \cite{DBLP:journals/corr/abs-2304-14178}, \textbf{LLaMA-Adapter V2} \cite{DBLP:journals/corr/abs-2304-15010}, \textbf{Multimodal-GPT} \cite{DBLP:journals/corr/abs-2305-04790}, \textbf{LaVIN-7B}, \textbf{LaVIN-13B} \cite{DBLP:journals/corr/abs-2305-15023}, \textbf{Lynx} \cite{DBLP:journals/corr/abs-2307-02469}, \textbf{BLIP-2} \cite{DBLP:journals/corr/abs-2301-12597}, and \textbf{InstructBLIP} \cite{DBLP:journals/corr/abs-2305-06500}.
Details of the different models are provided in Table \ref{Table_different_models} of Appendix \ref{Appendix_large_models}. A detailed description of the various LMs and the specific models utilized in our evaluation can also be found in the Appendix \ref{Appendix_LLMs} and \ref{Appendix_MLLMs}.

\subsection{Metrics}
Comprehensive metrics
are introduced to evaluate the performance of different models ($\mathcal{M}$) and various instructions ($\mathcal{I}$) on multiple datasets ($\mathcal{D}$), encompassing diverse multimodal content comprehension tasks.
We choose Accuracy, denoted as \textbf{acc$_{mdi}$}, as our primary metric to assess the performance of one LM, $m \in \mathcal{M}$, with the designed multimodal instruction, $i \in \mathcal{I}$, on each dataset, $d \in \mathcal{D}$.
\begin{equation}
    \begin{aligned}
        p^j_{mdi} = LM(T^j,V^j), \  acc_{mdi} = \frac{\sum_{j=1}^{N_d} {\mathbbm{1}}_{(p^j_{mdi}=l^j_{mdi})}} {N_d},
    \end{aligned}
\end{equation}
where LM represents a specific language model, the LLM takes only the text instruction, including text context, as input, denoted as $T$, while the MLLM takes the multimodal content, $(T, V)$, as input. 
For the $j$-th instance, $p^j_{mdi}$ represents the predicted label of the LM, $l^j_{mdi}$ is the true label. 
$\mathbbm{1}$ means the indicator function, $N_d$ is the number of instances, $acc_{mdi} \in Acc_{mdi}$ is the accuracy performance, $Acc_{mdi} \in \mathbb{R}^{|\mathcal{M}| \times |\mathcal{D}| \times |\mathcal{I}|}$, $|*|$ means the number of $*$. 
$|\mathcal{M}|=20$, $|\mathcal{D}|=14$, and $|\mathcal{I}|=10$ represent the number of evaluated models, datasets, and instructions, respectively.
\paragraph{Best Performance}
Considering the performance variations across different instructions, we report the best performance, \textbf{best accuracy} \textbf{$A^{\tilde{i}} \in \mathbb{R}^{|\mathcal{M}| \times |\mathcal{D}|}$}, achieved by each model among all instructions on each dataset. This metric highlights the upper bounds of performance for different models.
\begin{equation}
    \begin{aligned}
    A_{md} = Max(\{(acc_{mdi})\}_{i \in \mathcal{I}}), \\
     \tilde{i}_{md} = \mathop{\arg\max}_{i}\ \ \ (\{(acc_{mdi})\}_{i \in \mathcal{I}}),
    \label{equ_best_acc}
    \end{aligned}
\end{equation}
We acquire the best accuracy $A^{\tilde{i}}_{md}$. 
\paragraph{Mean Relative Gain (MRG)}
Given the diversity of models and instructions, it's understandable that we observe substantial variations in accuracy for each dataset, contingent on different models and various instructions.
Therefore, we leverage aggregating metrics to evaluate the overall performance across instructions and  models.
Following \cite{DBLP:journals/corr/abs-2307-00259}, on one hand, we calculate the \textbf{Mean Relative Gains of Models} (\textbf{MRG$^{\mathcal{M}}$}) to meaningfully compare and aggregate their performance across all instructions. 
On the other hand, we calculate the \textbf{Mean Relative Gains of Instructions}  (\textbf{MRG$^{\mathcal{I}}$}) to meaningfully compare and summarize their performance across all models.
We first calculate the mean relative gain for each model (MRG$^{\mathcal{M}}$). MRG$_{md}^{\mathcal{M}}$ indicates the percentage by which the model $m$ surpasses the average performance (across $\mathcal{M}$) on dataset $d$, when averaged across instructions $\mathcal{I}$.
\begin{equation}
  \begin{aligned}
        MRG^{\mathcal{M}}_{md}  = \frac{1}{|\mathcal{I}|} \sum_{i \in \mathcal{I}} r_{mdi}, MRG^\mathcal{M} \in \mathbb{R}^{|\mathcal{M}| \times |\mathcal{D}|}, \\
      r_{mdi}  = \frac{acc_{mdi} -\overline{Acc_{di}}}{\overline{Acc_{di}}} \times 100, \\
      \overline{Acc_{di}}  = \frac{1}{|\mathcal{M}|} \sum_{m \in \mathcal{M}} Acc_{mdi}, \overline{Acc_{di}} \in \mathbb{R}^{|\mathcal{D}| \times |\mathcal{I}|}. \\    
    \end{aligned}
    \label{Eq_model_MRG}
\end{equation}

We also calculate the mean relative gain for each instruction (MRG$^{\mathcal{I}}$). 
\textbf{MRG}$_{id}^{\mathcal{I}}$ represents the percentage value of the performance of instruction $i$ on dataset $d$ that is better than the average performance (calculated across all instructions in $\mathcal{I}$), considering all models in $\mathcal{M}$.
\begin{equation}
  \begin{aligned}
        MRG_{id}^{\mathcal{I}}  = \frac{1}{|\mathcal{M}|} \sum_{m \in \mathcal{M}} r_{idm}, MRG^\mathcal{I} \in \mathbb{R}^{|\mathcal{I}| \times |\mathcal{D}|}, \\
      r_{idm}  = \frac{acc_{idm} -\overline{Acc_{dm}}}{\overline{Acc_{dm}}} \times 100, \\
      \overline{Acc_{dm}}  = \frac{1}{|\mathcal{I}|} \sum_{i \in \mathcal{I}} Acc_{idm}, \overline{Acc_{dm}} \in \mathbb{R}^{|\mathcal{D}| \times |\mathcal{M}|}, \\
    \end{aligned}
    \label{Eq_Instruction_MRG}
\end{equation}
where the $Acc_{idm}$ is the transpose of $Acc_{mdi}$.

\begin{table*}[h] 
\small
\begin{center}
\caption{
The best zero-shot performance, $A^{\tilde{i}}$ ($\uparrow$, measured by Accuracy), of various large language models on different datasets, with superscripts indicating the specific instruction 
that resulted in the best performance for each dataset on the respective model. The ``\textbf{Total}" column represents the sum of accuracy scores across all datasets for each model.MVSA-S, MVSA-M, Twitter15, and Twitter17 refer to the MVSA-Single, MVSA-Multiple, Twitter-2015, and Twitter-2017 datasets, respectively.
AdapterV2 and MultiGPT represent LLaMA-AdapterV2 and Multimodal-GPT models, separately.
}
\vspace{-1em}
\resizebox{0.9\textwidth}{!}{
\begin{tabular}{
p{1.7cm}< \centering| 
p{1.2cm}< \centering |
p{1.2cm}< \centering p{1.2cm}< \centering p{1cm}< \centering| 
p{1cm}< \centering p{1cm}< \centering p{1cm}< \centering|
p{1cm}< \centering| p{1cm}< \centering| p{1cm}< \centering|
p{1cm}< \centering
}
\toprule[1pt]
\multirow{2}{*}{\textbf{Models}}
& \textbf{VQA}
& \multicolumn{3}{c|}{\textbf{MSA}} 
& \multicolumn{3}{c|}{\textbf{MABSA}}
& \textbf{MHMR} 
& \textbf{MSR}
& \textbf{MRE}  
& \multirow{2}{*}{\textbf{Total}} \\
\cline{2-11}
& \textbf{ScienceQA}
& \textbf{MVSA-S} & \textbf{MVSA-M} & \textbf{TumEmo} 
& \textbf{Twitter15} & \textbf{Twitter17} & \textbf{MASAD} 
& \textbf{Hate} 
& \textbf{Sarcasm} 
& \textbf{MNRE} 
\\
\midrule[1pt]
\textbf{ChatGPT} & 69.41$^7$ & 56.55$^3$ & 53.18$^7$ & 48.17$^3$ & 65.48$^4$ & 59.97$^{10}$ & 72.70$^3$ & 60.84$^8$ & 69.02$^7$ & \textbf{38.28$^2$} &  593.60\\
\textbf{LLaMA1-7B} & 36.19$^5$ & 67.23$^1$ & 60.72$^1$ & 38.26$^1$ & 58.53$^3$ & 46.43$^3$ & 65.67$^7$ & 50.40$^4$ & 58.99$^4$ & 2.66$^3$ &  485.09 \\
\textbf{LLaMA1-13B} & 43.33$^6$ & 66.99$^2$ & 68.82$^8$ & 44.68$^6$ & 52.07$^3$ & 47.24$^7$ & 65.49$^2$ & 49.2$^5$ & 57.53$^4$ & 19.22$^5$ &  514.59\\
\textbf{LLaMA2-7B} & 43.08$^6$ & 66.99$^1$ & 69.22$^2$ & 40.28$^4$ & 58.53$^3$ & 46.60$^2$ & 67.19$^2$ &	52.00$^1$ & 56.33$^1$ & 3.59$^7$ &  503.82\\
\textbf{LLaMA2-13B} & 55.78$^7$ & 66.02$^1$ & 68.69$^7$ & 45.78$^6$ & 60.37$^5$ & 48.54$^5$ & 69.10$^2$ & 55.00$^2$ & 60.23$^1$ & 20.00$^5$ &  549.50\\
\textbf{Flan-T5-XXL} & 67.43$^9$ & 64.81$^4$ & 66.01$^4$ & 49.56$^3$ & \textbf{72.13$^5$} & 63.70$^3$ & 74.39$^8$ & 57.4$^7$ & 71.40$^9$ & 31.41$^9$ &  618.23\\
\midrule[1pt]
\textbf{OpenFlamingo} &  39.27$^5$ & 55.58$^7$ & 61.15$^7$ & 29.47$^9$ & 57.28$^5$ & 46.19$^5$ & 66.91$^7$ &	49.40$^2$ & 52.68$^1$ & 3.12$^6$ &  461.06\\
\textbf{Fromage} & 34.51$^7$ & 29.85$^6$ & 28.19$^2$ & 22.76$^1$ & 19.96$^6$ & 27.31$^6$ & 35.10$^6$ &	37.60$^2$ & 40.68$^7$ & 0.16$^1$ &  276.12\\
\textbf{LLaVA-7B} & 41.10$^5$  & 56.55$^7$ & 64.71$^3$ & 4.26$^1$ & 28.26$^7$ & 37.28$^7$ & 58.22$^7$ &	22.80$^8$ & 13.45$^8$ & 0 &  326.62\\
\textbf{LLaVA-13B} & 47.74$^7$  & 58.01$^3$ & 66.40$^3$ & 5.04$^7$ & 28.35$^3$ & 36.87$^3$ & 59.84$^5$ &	28.40$^4$ & 28.73$^4$ & 0 &  359.39\\
\textbf{MiniGPT4} & 58.70$^5$ & 71.12$^5$ & 69.54$^5$ & 49.28$^2$ & 37.99$^{5,6}$ & 48.32$^{5,6}$ & 65.71$^{5,6}$ &	48.40$^8$ & 38.40$^7$ & 2.81$^6$ & 490.27\\
\textbf{mPLUG-Owl} & 37.93$^7$ & 53.88$^1$ & 46.98$^1$ & 34.74$^1$ & 36.35$^2$ & 40.36$^2$ & 59.27$^2$ & 49.20$^7$ & 39.77$^7$ & 8.75$^5$ &  407.22\\
\textbf{AdapterV2} & 54.44$^7$ & 59.95$^{3,5}$ & 68.63$^3$ & 5.56$^8$ & 30.57$^5$ & 39.79$^5$ & 64.07$^5$ & 49.00$^{2,4}$ & 39.77$^4$ & 0.16$^7$ &  411.93 \\
\textbf{VPGTrans} & 47.00$^3$ & 59.22$^7$ & 68.43$^7$ & 8.01$^7$ & 28.35$^5$ & 33.39$^3$ & 61.88$^3$ &	45.20$^5$ & 32.46$^5$ & 0.31$^3$ & 384.26 \\
\textbf{MultiGPT} & 36.29$^5$ & 52.91$^7$ & 62.03$^1$ & 30.26$^2$ & 58.53$^5$ & 46.35$^5$ & 67.58$^7$ &	49.80$^2$ & 59.82$^4$ & 2.81$^1$ &  466.38\\
\textbf{LaVIN-7B} & 75.11$^3$ & 39.32$^2$ & 40.75$^2$ & 26.84$^7$ & 37.22$^1$ & 33.06$^1$ & 60.08$^7$ & 50.40$^7$ & 60.48$^7$ & 12.34$^5$ &  435.62\\
\textbf{LaVIN-13B} & \textbf{77.54$^5$}& 53.64$^4$ & 48.79$^4$ & 32.77$^4$ & 35.39$^6$ & 40.68$^6$ & 62.76$^6$ &	49.60$^1$ & 57.58$^7$ & 11.56$^1$ &   470.31\\
\textbf{Lynx} & 38.28$^7$ & 64.32$^7$ & 67.71$^9$ & 42.79$^6$ & 46.00$^6$ & 47.00$^{10}$ & 73.52$^2$ &	51.60$^7$ & 43.96$^7$ & 9.22$^9$ &  484.39\\
\midrule[1pt]
\textbf{BLIP-2} & 74.17$^1$  & 66.26$^4$ & 68.22$^9$ & 51.06$^3$ & 70.78$^5$ & \textbf{64.42$^3$} & 77.59$^9$ &	58.00$^7$ & 72.02$^2$ & 34.69$^1$ &   637.21 \\
\textbf{InstructBLIP} & 73.33$^2$  & \textbf{71.60$^6$} & \textbf{70.37$^6$} & \textbf{52.36$^8$} & 63.07$^5$ & 62.72$^3$ & \textbf{80.53$^{10}$} &	\textbf{58.20$^9$} & \textbf{73.10$^7$} & 36.72$^2$ & \textbf{642.00 }\\
\bottomrule[1pt]
\end{tabular}
}
\label{Best_Acc}
\end{center}
\vspace{-1em}
\end{table*}
\paragraph{Stability}
Stability is another crucial indicator of both model and instruction. We measure the stability of a model (given various instructions) using \textbf{Model Stability Metrics},  \textbf{$S^{\mathcal{M^{'}}}$}. This metric is determined by calculating the standard deviation of the accuracy of each model when varying the instructions for each dataset.
Similarly, we assess the stability of an instruction (given different LMs) using \textbf{Instruction Stability Metrics}, \textbf{$S^{\mathcal{I^{'}}}$}. This metric is obtained by computing the standard deviation of the accuracy under each instruction across different models for each dataset.
Note that a few models may consistently perform poorly across all instructions, which results in good stability, albeit in a negative sense. Therefore, we evaluate the stability of LMs and instructions that perform well.
\begin{equation}
    \begin{aligned}
        S^{\mathcal{M}^{'}}_{md} = \sqrt{\frac{1}{|\mathcal{I}|} \sum_{i \in \mathcal{I}} (acc_{mdi} - \overline{Acc_{md}})^2}.
    \end{aligned}
    \label{Eq_model_stability}
\end{equation}
\begin{equation}
    \begin{aligned}
        S^{\mathcal{I}^{'}}_{id} = \sqrt{\frac{1}{|\mathcal{M}|} \sum_{m \in \mathcal{M}} (acc_{idm} - \overline{Acc_{id}})^2},
    \end{aligned}
    \label{Eq_Instruction_stability}
\end{equation}
where $\overline{Acc_{md}}$ and $\overline{Acc_{id}}$ are the transpose of $\overline{Acc_{dm}}$ and $\overline{Acc_{di}}$, respectively. $\mathcal{M}^{'} \subset \mathcal{M}$, and $\mathcal{I}^{'} \subset \mathcal{I}$. 
\begin{table*}[t]  \small
\begin{center}
\caption{The mean relative gain, MRG$^{\mathcal{M}}$ ($\uparrow$), for various models across instructions, .
Positive values indicate performance above the average, while negative values indicate performance below the average. The `Wins1' and `Wins3' columns display the number of tasks where a model achieves the highest aggregated performance and the top 3 aggregated performance, respectively.
Text in `\textbf{bold}', `\underline{underline}', and `$\star$'  indicate the best scores, the sub-optimal scores, and the third-best scores, respectively.
}
\vspace{-1em}
\resizebox{0.9\textwidth}{!}{
\begin{tabular}{
p{1.8cm}< \centering| 
p{1.2cm}< \centering |
p{1.1cm}< \centering p{1.2cm}< \centering p{1cm}< \centering|
p{1cm}< \centering p{1cm}< \centering p{1cm}< \centering|
p{0.9cm}< \centering| p{0.9cm}< \centering| p{0.9cm}< \centering|
p{0.5cm}< \centering p{0.6cm}< \centering
}
\toprule[1pt]
\textbf{Models} 
& \textbf{ScienceQA}
& \textbf{MVSA-S} & \textbf{MVSA-M} & \textbf{TumEmo} 
& \textbf{Twitter15} & \textbf{Twitter17} & \textbf{MASAD} & 
\textbf{Hate} & \textbf{Sarcasm} & \textbf{MNRE} & 
\textbf{Wins1} & \textbf{Wins3}\\
\midrule[1pt]
\textbf{ChatGPT} & 71.44 & 9.52  & 1.31  &  86.03 & \textbf{127.92} & \underline{82.85} &  56.38 &  59.03 &  69.78 &  291.99 & \textbf{1} & \textbf{2}\\
\textbf{LLaMA1-7B} & -45.26 &  -43.08 &  -46.98 &  -13.85 &  -27.61 &  -33.06 &  -39.77 &    5.89 &   -3.63   & -90.92 & 0 & 0 \\
\textbf{LLaMA1-13B} & -47.07 &   -9.12 &  -6.95 & -13.85 &  -3.90 & 1.02   & -5.31  &  -2.73 &  -14.69
  &  -46.82 & 0 & 0\\
\textbf{LLaMA2-7B} & -30.91 &   -0.25 &   -2.06 &    2.49 &  -26.8 &   -21.70 &   -28.65 &  -13.30 &     2.49 & -83.94 & 0 & 0 \\
\textbf{LLaMA2-13B} & -42.92 & -15.03 & -18.04  & 14.4 &  -23.81 & -25.09 & -37.27 & -16.71 & -12.88 &
-63.5 & 0 & 0\\
\textbf{Flan-T5-XXL} &  84.90$^\star$   &  43.91$^\star$   & 38.29  & 105.05$^\star$  &  \underline{88.97}   & 84.50  & 64.30$^\star$  &   61.19$^\star$   & 96.37$^\star$ &  293.38$^\star$  & 0 & \textbf{8}\\
\midrule[1pt]
\textbf{OpenFlamingo} & -54.56  & -48.87 &  -46.66 &  -34.36 &  -44.16 &  -46.32 &  -45.13 &  -30.41  & -36.56 & 87.95  & 0 & 0\\
\textbf{Fromage} & -66.77 & -72.14 & -84.82 & -75.84 & -80.04  & -77.48 & -72.30  & -51.60 &  -43.94 & -99.19  & 0 & 0\\
\textbf{LLaVA-7B} &  2.34  & 16.74 &  35.95 & -86.19 & -15.08 &  -1.25 &  12.26 & -59.73 & -70.09 &
  -100.00   & 0 & 0 \\
\textbf{LLaVA-13B} &  9.28 &  28.69  & 44.76 & -82.93 &  -0.73  & 13.01  & 27.03 & -46.13 & -48.23 &
  -100.00  & 0 & 0 \\
\textbf{MiniGPT4} &  28.49 &  26.32 &  20.42  & 92.87 & -16.18  & -2.00  &  -34.05  &  9.51 & -34.58 &
 -80.11 & 0 & 0\\
\textbf{mPLUG-Owl} & -31.52&   -3.15 &  -20.31 &    2.83 &  -23.44 &  -24.88 &  -12.35 &  10.96  & -7.06
  & -47.16  & 0 & 0\\
\textbf{AdapterV2} &  36.87 &  27.03  & \underline{52.74}  & -83.46 &  -9.42 &   4.99 &   8.04 &  -2.44 & -33.24 & -99.81 & 0 & \textbf{1} \\
\textbf{VPGTrans} & -19.50  &   5.92 &  20.93 & -77.59 & -38.53 & -34.34 & -19.54 & -44.01 & -52.42 & -98.55 & 0 & 0 \\
\textbf{MultiGPT} & -56.37 & -47.94 & -43.88 & -34.58 & -32.77 & -38.77 & -30.05 & -14.55 & -21.10 &
-87.89 & 0 & 0 \\
\textbf{LaVIN-7B} &  3.05 & -34.29 & -35.51 & -37.24 & -32.83 & -40.02 & -13.34 &  12.28 &  37.33 & -68.83 & 0 & 0 \\
\textbf{LaVIN-13B} & 7.97  &-21.92 & -39.21 &  -5.31 & -24.72 & -22.27 &  0.68 &   0.11 &  -4.72 & -38.64  & 0 & 0\\
\textbf{Lynx} & -45.75 &  15.88 &  19.05  &  7.3  &  14.78 &  14.38  & 23.39  & -4.67 & -24.52 & -50.31  & 0 & 0\\
\midrule[1pt]
\textbf{BLIP-2} & \underline{94.94}  & \underline{55.19} &  51.72$^\star$ & \underline{111.78} &  85.9$^\star$  &  82.77$^\star$ &  \underline{70.35} &  \underline{62.95} &  \underline{99.05} & \underline{319.07}  & 0 & \textbf{10}\\
\textbf{InstructBLIP} & \textbf{101.33} &  \textbf{66.59}  & \textbf{59.26} & \textbf{122.45} &  82.45  & \textbf{83.65}  & \textbf{75.32} &  \textbf{64.37} & \textbf{102.66} & \textbf{339.17}  & \textbf{9} & \textbf{9}\\
\bottomrule[1pt]
\end{tabular}
}

\label{Table_model_mean_relative_gain}
\end{center}
\vspace{-1em}
\end{table*}
\paragraph{Adaptability}
Different instructions have a significant impact on model performance. To quantify the adaptability between LMs and instructions, we propose the \textbf{Global Top-K Hit Ratio}, $GHR@K$, as a metric to evaluate the performance of each instruction on different models. This metric measures the proportion of times each instruction achieves top-K performance on a specific model across all datasets. We aim to find out which specific instructions for each model have excellent performance across all dataset. As such, we evaluate each instruction based on its overall Top-K Hit Ratio score across all datasets for the $m$ model.
\begin{equation}
    \begin{aligned}
     I^K_{d|m} = \mathop{\arg\max}_{i^1, ...  , i^K}K\ \ \ (\{(acc_{mdi})\}_{i \in \mathcal{I}}),  |I^K_{d|m}| = K, \\
    GHR@K_{m} = \frac{Counter_m^{\mathcal{I}}([I^K_{d^1|m} || ... || I^K_{d^{|\mathcal{D}|} |m}])}{(|\mathcal{D}| \times K)},
    \label{Eq_hit_ratio}
    \end{aligned}
\end{equation}
where $K=3$ represents the number of instructions that show the top-K performance,
$||$ is the contact operation, and
``Counter$_m^{\mathcal{I}}$'' is a dictionary function that counts the number of occurrences of each instruction $i \in \mathcal{I}$ for model $m$. $|GHR@K_{m}| = |\mathcal{I}|$.

\section{Experimental Results and Analysis}
We evaluate the zero-shot performance of 20 LLMs on 14 datasets, with each assessment using 10 instructions, as outlined in Table \ref{Table_dataset_statistic}, Figure \ref{Figure_1_ScienceQA_different_instructions}, \ref{Figure_1_MSA_Multimodal_Instruction}, and \ref{Figure_2_text_instruction_for_different_tasks}. Consequently, we gather a total of 2800 (20 $\times$ 14 $\times$ 10) experimental results, denoted as $Acc_{mdi}$.
Except for ScienceQA, other datasets primarily depend on the text modality for inference. Therefore, we conduct evaluations using a series of popular text-only LMs (Part 1 of Table \ref{Best_Acc}). While ScienceQA primarily relies on image modalities to answer questions, we also aim to extract potential knowledge from models solely through text modalities. As illustrated in Figure \ref{Figure_1_ScienceQA_different_instructions}, answers can be provided solely based on the text input without the necessity of the image modality. 
Note tha in contrast to other multimodal tasks involving text-image pairs, the original MOSI and MOSEI datasets are video-based. We preprocess these two datasets into the text-image format for our evaluation. Experimental results for the MOSI and MOSEI datasets are available in Appendix \ref{Appendix_Results_MOSI_MOSEI}, while results for other datasets are presented in the main paper.
Even though we design only 10 manual instructions to assess the performance of various models across different datasets, these evaluations can provide valuable insights, enabling us to draw meaningful conclusions. The specific analyses are given in the subsequent sections.

\begin{table*}[!thp] \small
\begin{center}
\caption{The mean relative gain, MRG$^{\mathcal{I}}$ ($\uparrow$), for different instructions (\# 1, ..., \# 10) across  models.
}
\vspace{-1em}
\resizebox{0.75\textwidth}{!}{
\begin{tabular}{
p{1.5cm}< \centering| 
p{1.2cm}< \centering |
p{1.1cm}< \centering p{1.2cm}< \centering p{1cm}< \centering|
p{1cm}< \centering p{1cm}< \centering p{1cm}< \centering|
p{0.9cm}< \centering| p{0.9cm}< \centering| p{0.9cm}< \centering|
p{0.5cm}< \centering p{0.6cm}< \centering
}
\toprule[1pt]
\textbf{Instructions} & \textbf{ScienceQA} & \textbf{MVSA-S} & \textbf{MVSA-M} & \textbf{TumEmo} & \textbf{Twitter15} & \textbf{Twitter17} &   \textbf{MASAD} & \textbf{Hate} & \textbf{Sarcasm} & \textbf{MNRE} &\textbf{Wins1} &\textbf{Wins3} \\
\midrule[1pt]
\textbf{\# 1} & -32.34 &  \underline{31.61} & \underline{25.82}  & \textbf{30.57}  & 7.66  & 20.12  & 14.55  & 12.11  & 24.29$^\star$ &  1.98 & \textbf{1} & \textbf{4}\\
\textbf{\# 2} & 14.66$^\star$ & \textbf{40.20} &   \textbf{38.24} &  \underline{30.24} &  21.95$^\star$ &  \textbf{31.20} &   \underline{32.49} &  \textbf{24.44} & 24.15 & -3.98 & \textbf{4} & \textbf{8}\\
\textbf{\# 3} &  9.26 & -27.06 & -19.3 &  -27.39 &  \underline{32.33} &  10.71 &  -5.34 &  -4.26 & -33.85 & 15.69$^\star$ & 0 & \textbf{2}\\
\textbf{\# 4} & -27.73 & -26.2 &  -23.08 &  -6.97 & -33.38 &  -33.66 & -35.05 &  -2.16 &  \underline{24.99} & -50.02 & 0 & \textbf{1}\\
\textbf{\# 5} & \textbf{46.66} &  -8.84 &  -1.41 & -16.25 & \textbf{41.25} &  \underline{23.00} &    10.96 &  12.66$^\star$ &  -7.56 & \underline{37.91} & \textbf{2} & \textbf{5}\\
\textbf{\# 6} & 6.48 & 13.63 &  14.97 &  15.52$^\star$ &  16.42  &  19.95 &  24.67$^\star$ & -15.2 & -1.06 & -16.27 & 0 & \textbf{2}\\
\textbf{\# 7} & \underline{42.88} & 28.34$^\star$ &  25.52$^\star$ &  15.43 &   17.13 &  22.78$^\star$ &  \textbf{42.88} &  \underline{21.77} &  \textbf{25.59} & \textbf{57.81} & \textbf{3} & \textbf{8}\\
\textbf{\# 8} & -38.18 &  -9.35 &  -9.46 &  -9.85 & -48.74 & -45.77 & -44.17 & -16.07 & -16.94 & -60.56& 0 & 0\\
\textbf{\# 9} & 8.82 & -16.38 & -18.18 &  -2.87 & -18.77 & -13.31 &  -6.18 & -15.96 & -22.09 & -35.03 & 0 & 0\\
\textbf{\# 10} & -30.53 & -25.95 & -33.12 & -28.42 & -35.84 & -35.02 & -34.82 & -17.34 & -17.52 & -47.53 & 0 & 0\\
\bottomrule[1pt]
\end{tabular}
}
\label{Table_instruction_mean_relative_gain}
\end{center}
\vspace{-1em}
\end{table*}
\begin{table*}[h] \small
\begin{center}
\renewcommand{\arraystretch}{1} 
\caption{The stability, $S^{\mathcal{M^{'}}}$ ($\downarrow$), of various models with excellent performance across instructions. 
}
\vspace{-1em}
\resizebox{0.75\textwidth}{!}{
\begin{tabular}{
p{1.8cm}< \centering| 
p{1.2cm}< \centering |
p{1.1cm}< \centering p{1.2cm}< \centering p{1cm}< \centering|
p{1cm}< \centering p{1cm}< \centering p{1cm}< \centering|
p{0.9cm}< \centering| p{0.9cm}< \centering| p{0.9cm}< \centering|
p{0.6cm}< \centering
}
\toprule[1pt]
\textbf{Models} & \textbf{ScienceQA}
& \textbf{MVSA-S} & \textbf{MVSA-M} & \textbf{TumEmo} & \textbf{Twitter15} & \textbf{Twitter17} & \textbf{MASAD} & 
\textbf{Hate} & \textbf{Sarcasm} & \textbf{MNRE} & 
\textbf{Wins1} \\
\midrule[1pt]
\textbf{ChatGPT} & 4.36 & 13.44 & 8.37 & 3.61 & \textbf{2.53} & 3.2 &  6.32 & 6.9 &  9.75 & 6.92 & \textbf{1}\\
\textbf{Flan-T5-XXL} & \textbf{0.57} &  10.84 &  8.99 &  4.62 &  9.63  & \textbf{3.09} &  \textbf{0.52} &  0.93 &  1.75  & 2.19 & \textbf{3}\\
\midrule[1pt]
\textbf{BLIP-2} & 5.73 &  5.46 &  7.86 &  3.05 & 9.04  & 3.55  & 0.53  & 0.69  & 1.52  & \textbf{1.89} & \textbf{1}\\
\textbf{InstructBLIP} & 0.73 &  \textbf{4.15} &  \textbf{4.44}  & \textbf{0.62} &  7.02  & 3.29  & 0.92 &  \textbf{0.65}  & \textbf{1.27}  & 3.90 & \textbf{5}\\
\bottomrule[1pt]
\end{tabular}
}
\label{stability of various models}
\end{center}
\vspace{-1em}
\end{table*}

\begin{table*}[!htp] \small
\begin{center}
\renewcommand{\arraystretch}{1} 
\caption{The stability, $S^{\mathcal{I^{'}}}$ ($\downarrow$), of different instructions with excellent performance across models.
}
\vspace{-1em}
\resizebox{0.75\textwidth}{!}{
\begin{tabular}{
p{1.8cm}< \centering| 
p{1.2cm}< \centering |
p{1.1cm}< \centering p{1.2cm}< \centering p{1cm}< \centering|
p{1cm}< \centering p{1cm}< \centering p{1cm}< \centering|
p{0.9cm}< \centering| p{0.9cm}< \centering| p{0.9cm}< \centering|
p{0.6cm}< \centering
}
\toprule[1pt]
\textbf{Instructions} & \textbf{ScienceQA} & \textbf{MVSA-S} & \textbf{MVSA-M} & \textbf{TumEmo} & \textbf{Twitter15} & \textbf{Twitter17} & \textbf{MASAD} & 
\textbf{Hate} & \textbf{Sarcasm} & \textbf{MNRE} & 
\textbf{Wins1} \\
\midrule[1pt]
\textbf{\# 1}  & 23.20 & 15.00 &   16.75 & \textbf{15.53} & 13.30 & 13.82 & 18.91 & 15.55 & 19.78 & 12.50 & \textbf{1}\\
\textbf{\# 2} & 17.36 & \textbf{12.38} & \textbf{16.32} & 15.57 & \textbf{11.83} & \textbf{11.89} & 15.09 & \textbf{15.10} &  18.74 & 13.41 & \textbf{5}\\
\textbf{\# 3} & 21.73 & 24.41 & 23.80 &  18.00 &   19.10 & 17.49 & 23.82 & 19.65 & 23.35 & \textbf{10.44} & \textbf{1}\\
\textbf{\# 4} & 23.88 & 23.66 & 26.06 & 17.27 & 18.10 & 19.79 & 28.14 & 17.47 & 19.96 & 12.72 & 0\\
\textbf{\# 5} & 16.22 & 21.30 &  21.01 & 18.56 & 20.61 & 17.20 &  19.51 & 16.36 & 21.15 &  11.67 & 0\\
\textbf{\# 6} & 17.35 & 15.52 & 18.12 & 17.16 & 13.33 & 14.62 & 17.25 & 17.91 & 20.61 & 11.34 & 0\\
\textbf{\# 7} & \textbf{14.01} & 14.58 & 18.18 & 16.33 & 12.63 & 13.03 & \textbf{11.94} & 15.45 & \textbf{18.98} & 11.49 & \textbf{3}\\
\bottomrule[1pt]
\end{tabular}
}
\label{stability of various instructions}
\end{center}
\vspace{-1em}
\end{table*}

\subsection{Best Performance}
We present the best performance ($A^{\tilde{i}}$), calculated using Eq. \ref{equ_best_acc}, achieved by different models across various multimodal datasets in Table \ref{Best_Acc} and Table \ref{Table_Best_Acc_MOSI_MOSEI} (Appendix \ref{Appendix_Best_Acc_MOSI_MOSEI}).
Thereby, we provide benchmarks for different models on each dataset and make the following observations:
(1) InstructBLIP excels by achieving the top performance on six datasets and secures the first position in the `total' metric, with BLIP-2 closely following in second place. It's worth noting that both InstructBLIP and BLIP2 use Flan-T5-XXL as their backbone model, and Flan-T5-XXL emerges as the best-performing large language model among the 6 LLMs (the first part in Table \ref{Table_different_models}). This clearly underscores the exceptional performance of Flan-T5-XXL. The reason for this is that, the LLaMA series models have a Decoder-only architecture, while Flan-T5-XXL is based on an Encoder-Decoder architecture. The encoder module of the latter provides significant advantages in multimodal representation learning for multimodal content comprehension tasks.
(2) Except for InstructBLIP and BLIP-2, the `total' metric of most MLLMs across all datasets is lower than that of pure large language models. We hypothesize that this phenomenon could be attributed to the possibility that underperforming MLLMs may not effectively achieve multimodal understanding in multimodal content comprehension tasks.
Compared to Flan-T5-XXL, InstructBLIP and BLIP-2 show improvement across most tasks through comprehensive vision-language representation learning.
(3) ScienceQA primarily answers questions based on the image modality. Therefore, we also conduct comparative experiments without text context, where there is no `Context' item in Figure \ref{Figure_1_ScienceQA_different_instructions}. The experimental results are displayed in Table \ref{Table_ScienceQA} of Appendix \ref{Appendix_Comparison_of_ScienceQA}. We find that most MLLMs performs better on ScienceQA with text context; the image modailty is more important for ScienceQA.
(4) For the same model, a model with a larger size, indicated by a higher number of parameters, such as LLaMA-7B and LLaMA-13B, LLaVA-7B and LLaVA-13B, LaVIN-7B and LaVIN-13B, tends to perform better on most datasets. This suggests that larger models with more extensive training have greater potential to achieve higher performance.
(5) On one hand, within the same model, different datasets attain their best performance with varying instructions. On the other hand, for a given dataset, different models achieve their highest performance when paired with specific instructions. Currently, there is limited research dedicated to the selection of suitable instructions for diverse datasets and models. This area remains largely unexplored and offers significant potential for future investigation.
(6) All models perform poorly on the MNRE task, with ChatGPT achieving the best performance. This is mainly attributed to the significant long-tail effect present in the 19 categories within MNRE, suggesting ample room for further exploration in LMs for MNRE.
(7) LaVIN models exhibit the best performance on the ScienceQA task, primarily because LaVIN has a dedicated adapter trained specifically for ScienceQA, as illustrated in Table \ref{Table_different_models}
.  However, it is unfair to other models.

\begin{table*}[t] \small
\begin{center}
\caption{The Top-K Hit Ratio, $GHR@K$ ($\uparrow$), for various instructions (\# 1, ..., \# 10) across different LLMs on all datasets including MOSI and MOSEI.
`LLMs-Total' represents the sum of Top-K Hit Ratio scores across all pure LLMs (6 models) for each instruction.
`MLLMs-Total' represents the scores across all MLLMs (14 models).
`Total' signifies the scores across all LMs (20 models).
}
\vspace{-1em}
\resizebox{0.8\textwidth}{!}{
\begin{tabular}{
p{2cm}< \centering| 
p{0.95cm}< \centering p{0.95cm}< \centering p{0.95cm}< \centering
p{0.95cm}< \centering p{0.95cm}< \centering p{0.95cm}< \centering
p{0.95cm}< \centering p{0.95cm}< \centering p{0.95cm}< \centering
p{0.95cm}< \centering |
p{1.2cm}< \centering 
}
\toprule[1pt]
\textbf{Models} & \textbf{\# 1} & \textbf{\# 2} & \textbf{\# 3} & \textbf{\# 4} & \textbf{\# 5} & \textbf{\# 6} & \textbf{\# 7} & \textbf{\# 8} & \textbf{\# 9} &\textbf{\# 10} & \textbf{Variance}\\
\midrule[1pt]
 
\textbf{ChatGPT} & 9.52 & 9.52 & 11.90 & 9.52 & \textbf{21.43} &  0 &   19.05 & 11.90 &   0  &    7.14 & 42.86\\
\textbf{LLaMA1-7B} & \textbf{26.19}  & 21.43  & 7.14  & 4.76 &  9.52 &  2.38 & 21.43 &  4.76 &  0  & 2.38 & 80.29 \\
\textbf{LLaMA1-13B} & \textbf{16.67} & 16.67 &  4.76 &  4.76 & 14.29 & 14.29 & 16.67 & 11.90 &   0 &  0 & 42.88 \\
\textbf{LLaMA2-7B} & \textbf{21.43} & \textbf{21.43} & 11.90 &   7.14 & 14.29 &  7.14  & 7.14 &  0 &    9.52 &  0 & 50.81  \\
\textbf{LLaMA2-13B} & 16.67 & \textbf{26.19} &  0 &    2.38 & 19.05 & 14.29 & 19.05 &  2.38 &  0 &    0 & 90.49  \\
\textbf{Flan-T5-XXL} &  4.76 &  9.52 &  9.52  & 9.52  & 9.52 & 11.90 &  11.90 &   4.76 & \textbf{14.29} & \textbf{14.29} & 9.99 \\
\midrule[0.5pt] 
\textbf{LLMs-Total} & \underline{95.24} & \textbf{104.76} & 45.22 & 38.08 & 88.10 & 50.00 & \underline{95.24} & 35.70 & 23.81 & 23.81 & -\\
\midrule[1pt]
\textbf{OpenFlamingo} & \textbf{26.19} & 21.43 &  2.38 &  0  &  11.90  &  2.38 & 21.43 &  4.76 &  9.52 &  0 & 87.08\\
\textbf{Fromage} & 11.90 &  \textbf{30.95} &  2.38 &  0   & 4.76 & 21.43 & 23.81 &  0  & 4.76  & 0 & 117.69 \\
\textbf{LLaVA-7B}  & 4.76 &  11.9 & 16.67 & 0 & \textbf{19.05} &   7.14 &  \textbf{19.05} &   7.14 &   9.52 &   4.76 & 38.34\\
\textbf{LLaVA-13B} & 2.38 &  9.52 & 14.29 & \textbf{16.67} & 14.29 &  9.52 & \textbf{16.67} &  2.38 & 11.90 & 2.38 & 30.4\\
\textbf{MiniGPT4} &  2.38 & \textbf{16.67} & 14.29 & 11.90  & \textbf{16.67} & 11.90  &  7.14 &  2.38 &  4.76 & 11.90 & 27.0\\
\textbf{mPLUG-Owl} & \textbf{26.19} & 19.05 &  2.38 &  0 &    4.76 & 14.29 & 16.67 &  2.38 & 14.29 &  0 & 76.89 \\
\textbf{AdapterV2} & \textbf{14.29} & 11.90 &  11.90 &   7.14 & 11.90 &   0 &    7.14 & \textbf{14.29} &  9.52 & 11.9 & 16.78\\
\textbf{VPGTrans} &  2.38 &  2.38 & \textbf{21.43}& 14.29 & 16.67 & 11.90 &  \textbf{21.43} &  0 &    7.14 &  2.38 &61.02 \\
\textbf{MultiGPT} & \textbf{23.81} & \textbf{23.81}  & 7.14 &  2.38 & 14.29 &  2.38 & 21.43  & 0 & 4.76  & 0  & 88.22\\
\textbf{LaVIN-7B} & 11.90 &  \textbf{19.05} & 11.90 &   7.14 & 16.67 &  4.76 & 16.67 &  2.38 &  2.38 &  7.14 & 33.80\\
\textbf{LaVIN-13B} & 14.29 & \textbf{21.43} &  4.76 & 14.29 &  9.52 & 11.90 &  \textbf{21.43} &  0 &    0 &    2.38 & 58.75\\
\textbf{Lynx} & 7.14 & \textbf{19.05} &  2.38 &  4.76 &  2.38 & 14.29 & 16.67 &  0 &   21.43 & 11.90 & 53.08\\
\textbf{BLIP-2} &  4.76 & 14.29 &  7.14 & 11.90 &   4.76 &  9.52 & 11.90 &   2.38 & \textbf{16.67} & \textbf{16.67} &  23.6\\
\textbf{InstructBLIP} & 7.14 & 14.29  & 9.52 &  7.14 & 14.29 &  4.76  & 4.76  & 7.14 & \textbf{16.67} & 14.29  & 17.94\\
\midrule[0.5pt]
\textbf{MLLMs-Total} & 159.51 & \textbf{235.72} & 128.56 & 97.61 & 161.91$^\star$ & 126.17 & \underline{226.20} & 45.23 & 133.32 & 85.70 & -\\
\midrule[1pt] 
\textbf{Total} & 254.75$^\star$ & \textbf{340.48} & 173.78 & 135.69 & 250.01 & 176.17 & \underline{321.44} & 80.93 & 157.13 & 109.51 & -\\
\bottomrule[1pt]
\end{tabular}
}
\label{Instrcution_hit_ratio}
\end{center}
\vspace{-1em}
\end{table*}
\subsection{Mean Relative Gain}
Although the best performance metric reflects the upper limit of each model's capabilities, a comprehensive assessment necessitates a broader perspective. To evaluate the comprehensive performance of each model, we compute the mean relative gain across all instructions for each model on each dataset using Eq. \ref{Eq_model_MRG}, and the results are presented in Table \ref{Table_model_mean_relative_gain} and Table \ref{Table_Model_Mean_Relative_Gain_MOSI_MOSEI} (Appendix \ref{Appendix_Mean_Relative_Gain_for_Various LMs}).
The `Wins1' and `Wins3' metrics represent the number of tasks where a model or an instruction achieved the highest aggregated performance and the top-3 aggregated performance, respectively. 
Like the best model performance metric, models with Flan-T5-XXL as the backbone demonstrate relatively strong performance, including InstructBLIP, BLIP-2, and Flan-T5-XXL. 
Similarly, we calculate the mean relative gain of different instructions across all models to assess the overall performance of each instruction using Eq. \ref{Eq_Instruction_MRG}, and the results are shown in Table \ref{Table_instruction_mean_relative_gain} and Table \ref{Table_Instructions_Mean_Relative_Gain_MOSI_MOSEI} (Appendix \ref{Appendix_Mean_Relative_Gain_for_Various_Instructions}).
We observe that the mean relative gain across all models shows better performance on Instructions \# 1, \# 2, \# 5, and \# 7.
Among these instructions, \# 5 and \# 7 are both derived from \# 2, which is a simple Question-Answer mode. The reason for their good performance may be that models have been trained on relevant QA tasks, so specific words like `Question' and `Answer' aid in enhancing the model's performance. Additionally, we notice that the inclusion of option terms in instructions \# 5 and \# 7 has a positive impact on the models, improving their performance on specific datasets. It suggests that the formats of option terms can also influence the performance of language models.
\subsection{Stability}
Stability is also an important metric for assessing models and instructions.
As mentioned above, we calculate the stability metrics exclusively for models that exhibit strong overall performance, including ChatGPT, Flan-T5-XXL, BLIP2, and InstructBLIP. Similarly, we assess the stability of instructions that achieve a top-three average relative gain on at least one dataset, specifically instructions \#1 through \#7.
We separately evaluate the stability of models (ChatGPT, Flan-T5-XXL, BLIP-2, and InstructBLIP) using Eq. \ref{Eq_model_stability} and instructions (from \# 1 to \# 7) using Eq. \ref{Eq_Instruction_stability}, focusing on those that demonstrates excellent aggregated performance. Related results are presented in Table \ref{stability of various models}, Table \ref{stability of various instructions}, Table \ref{Table_stability_of_various_large_models_on_MOSI_MOSEI} (Appendix \ref{Appendix_the_stability_LMs}), and Table \ref{Table_stability_of various_instructions_on_MOSI_MOSEI} (Appendix \ref{Appendix_the_stability_instructions}). 
In terms of model stability, InstructBLIP, which also exhibits the best performance, demonstrates the highest stability. Compared to BLIP2, InstructBLIP goes a step further by applying instruction tuning to improve the stability of the model across various instructions. It demonstrates that instruction tuning can mitigate the sensitivity of models to different instructions in multimodal content comprehension tasks.
As for instruction stability, instruction \# 2 stands out with the highest stability, and it also showcases strong aggregated performance on this instruction. 
\subsection{Adaptability}
Since models are trained with distinct pre-training settings, they often display varying inclinations toward different instructions. We compute the adaptability between models and instructions using the Global Top-K Hit Ratio, as defined in Eq. \ref{Eq_hit_ratio} to quantify these preferences. The results, encompassing all datasets, including MOSI and MOSEI, are presented in Table \ref{Instrcution_hit_ratio}.
We advert that different models tend to perform better with specific instructions. Instruction \# 2 demonstrates the highest adaptability for all models, closely followed by \# 7. This phenomenon is consistent among both pure LLMs and MLLMs. It further demonstrates that most LMs are better suited for instructions designed in a Question-answer format.
We further find that models exhibit specific preferences for certain instructions. 
For instance, VPGTrans perfers Instruction \# 3 and \# 7 that use the optional term; 
LLaVA-7B, LLaVA-13B, and LaVIN-13B perform better on the instruction \#7, \# 5; moodels with Flan-T5-XXL as the backbone achieve better performance on Instruction \# 9 and \# 10;
while other models show excellent performance with Instruction \# 1 and \# 2.
It confirms our hypothesis that different models favor specific instructions, highlighting the significant influence of instruction design on model performance.
The LLaMA series models have a greater dependency on instructions, leading to larger performance disparities across different instructions and consequently, higher variance.
However, Flan-T5-XXL exhibits the lowest variance, signifying minimal performance differences across all instructions. Additionally, InstructBLIP and BLIP2, built upon Flan-T5-XXL, benefit from its characteristics, displaying low sensitivity to varying instructions.
We observe an interesting phenomenon where OpenFlamingo and Fromage across different instructions have high variance. It suggests that these models exhibit significant performance fluctuations across different instructions.
The primary reason for this variability is that OpenFlamingo and Fromage models have not undergone instruction tuning. In comparison to MLLMs that have been fine-tuned with instruction data, they are more sensitive to different instructions. In other words, MLLMs that have undergone instruction tuning perform better across various instructions, leading to more stable performance.
\section{Conclusion}
We conduct comprehensive evaluations and establish benchmarks for 20 LMs using 10 instructions, including 14 popular MLLMs, across 6 diverse multimodal content comprehension tasks. Our approach involves introducing a range of multi-perspective metrics for the comprehensive assessment framework, \textbf{\textsc{MM-BigBench}}. These metrics encompass the Best Performance metric, the Mean Relative Gain metric, the Stability metric, and the Adaptability metric.
We perform evaluation of MLLMs across various multimodal content comprehension tasks and draw important conclusions, including:
(1) Models with Flan-T5-XXL as the backbone, based on the Encoder-Decoder architecture, outperform the LLaMA series models with the Decoder-only architecture on our evaluated multimodal content comprehension tasks.
(2) Instructions in a `Question-Answer' format yield better performance.
(3) The performance of MLLMs trained using instruction tuning demonstrates increased stability across various tasks.
Our paper paves the way for new directions in further exploration within this rapidly evolving field.

\bibliographystyle{ACM-Reference-Format}
\bibliography{ME}

\appendix
\section{Related Works}
\label{Appendix_related_works}
\subsection{LLMs-Based Evaluation}
As Large Language Models (LLMs) \cite{DBLP:journals/corr/abs-2306-13549} gain popularity, numerous evaluations of LLMs have emerged.
For instance, \citet{DBLP:journals/corr/abs-2305-15005} conduct the evaluation of LLMs, including ChatGPT and FlanT5, across various sentiment analysis tasks, encompassing text sentiment analysis, aspect-based sentiment analysis, and multifaceted analysis of subjective texts. Additionally, \citet{DBLP:journals/corr/abs-2304-04339} assess the performance of ChatGPT in five representative sentiment analysis tasks.
Most of the studies mentioned above primarily concentrate on text sentiment analysis and assess only a limited number of LLMs. However, there is an increasing need for comprehensive evaluation frameworks.
LLMeBench introduces an open-source, user-friendly, and adaptable comprehensive benchmarking framework for LLMs. It incorporates four fundamental modules: the Dataset module, Asset module, Model module, and Evaluation module.
INSTRUCTEVAL \cite{DBLP:journals/corr/abs-2306-04757} offers a more extensive evaluation suite specifically designed for 11 instruction-tuned large language models. These include models like Flan-T5, Vicuna, Alpaca, and more. Additionally, InstructEval \cite{DBLP:journals/corr/abs-2307-00259} systematically investigates the generalizability of popular instruction selection and induction methods for in-context learning (ICL) in large language models (LLMs).

These studies primarily focus on evaluating LLMs and leave the performance of various Multimodal Large Language Models (MLLMs) in tasks related to multimodal content comprehension, which rely on both text and image modalities, largely unexplored. Furthermore, the work mentioned above either solely concentrates on assessing different models or exclusively evaluates the performance of instructions, neglecting the aspect of adaptability between models and instructions. Our primary focus is to assess the performance of various Multimodal Large Language Models on different multimodal content comprehension tasks and propose the Global Top-K Hit Ratio metric to quantify the adaptability between different models and various instructions.

\subsection{MLLMs-Based Evaluation}
Multimodal Large Language Models (MLLMs) \cite{DBLP:journals/corr/abs-2306-13549}, building upon the impressive performance of large language models, excel in a wide range of multimodal tasks, including Caption Generation, Visual Question Answering, and more. This leads to a surge in research focused on evaluating these models. For example, MME \cite{DBLP:journals/corr/abs-2306-13394} introduces the first MLLM evaluation benchmark, encompassing Perception Tasks (Recognition tasks and OCR) and Cognition Tasks (Commonsense Reasoning, Numerical Calculation, Text Translation, and Code Reasoning). Additionally, SEED-Bench \cite{DBLP:journals/corr/abs-2307-16125} evaluates 19K multiple-choice questions across 12 evaluation dimensions, covering comprehension of both image and video modalities. LVLM-eHub \cite{DBLP:journals/corr/abs-2306-09265} assesses 8 MLLMs, including InstructBLIP and MiniGPT, on 47 standard text-related visual benchmarks through quantitative capability evaluations and an online arena platform. Furthermore, MM-Vet \cite{DBLP:journals/corr/abs-2308-02490} proposes a benchmark for evaluating MLLMs on 16 tasks, defining six core Visual Language (VL) capabilities, including recognition, OCR, knowledge, language generation, spatial awareness, and math, to address complex multimodal tasks.

The previous evaluations of MLLMs primarily focus on image-driven text-related tasks to assess the conventional language-visual multimodal reasoning capabilities of the model. However, multimodal content comprehension tasks require a deeper understanding of multimodal content. Therefore, our work can be considered a complement to the existing studies mentioned above. When combined, these evaluations provide a more comprehensive and holistic assessment of MLLMs.

\section{Datasets}
\label{Appendex_datasets}
We conduct comprehensive evaluation of various Language Models across a diverse range of multimodal content comprehension tasks, including MSA, MABSA, MHMR, MSR, MRE, and VQA. Detailed statistics for each task and the datasets can be found in Table \ref{Table_dataset_statistic}. We further offer the detailed introduction to various multimodal content comprehension tasks and their corresponding datasets.
\subsection{Multimodal Sentiment Analysis}

Multimodal Sentiment Analysis (MSA) aims to detect the overall sentiment of a text-image pair or a video\cite{DBLP:conf/sigir/XuMC18, DBLP:journals/tmm/YangFW021, DBLP:conf/acl/YangF0W20, DBLP:conf/naacl/LiXZZ22, Yang2023MultipleCL}.
Our evaluation encompasses widely used three text-image datasets, including \textbf{MVSA-Single} and \textbf{MVSA-Multiple} datasets \cite{DBLP:conf/mmm/NiuZPE16}, as well as the \textbf{TumEmo} dataset \cite{DBLP:journals/tmm/YangFW021}; four video datasets, including \textbf{MOSI-2}, \textbf{MOSI-7} \cite{DBLP:journals/corr/ZadehZPM16}, \textbf{MOSEI-2}, and \textbf{MOSEI-7} \cite{DBLP:conf/acl/MorencyCPLZ18}. Considering that most MLLMs primarily accept text-image pairs as multimodal inputs, for video datasets, we initially extract one frame per second to create a candidate frame set. Subsequently, we randomly select only one\footnote{We also experiment with randomly selecting multiple frames, for example, three frames, and then inputting them individually into the MLLMs. We apply a voting principle to determine the final result. However, the outcomes are essentially comparable to randomly choosing a single frame. Consequently, we ultimately decide to randomly select one frame as the visual input for each video.} frame to serve as the image input for MLLMs. For MOSI-2 and MOSEI-2, the label space consists of \{positive, negative\}, with the exception of neutral, which is labeled as zero. In the case of MOSI-7 and MOSEI-7, the label space includes \{strongly positive, positive, weakly positive, neutral, weakly negative, negative, strongly negative\}, with neutral sentiment also being accounted for.
\begin{table*}[t] \small
\begin{center}
\renewcommand{\arraystretch}{1} 
\caption{Comprehensive Summary of Various Models. The abbreviations used in the table are as follows: `\textbf{PLLMs}' refers to the pretrained Language Models (LLMs) backbone of the Multimodal Large Language Models (MLLMs), `\textbf{PVM}' signifies the pretrained visual model backbone of the MLLMs, `\textbf{To-Paras}' and `\textbf{Tr-Paras}' represent the total number of parameters and trainable parameters for each language model. 
"\textbf{Held-In}" refers to the corresponding dataset trained or fine-tuned on the specific MLLM.
The `\textbf{GPU}' column indicates the single GPU utilization during inference, and the `\textbf{Time}' column signifies the time taken for model inference on each text/multimodal instance using single GPU. Note that GPU usage and inference time may vary slightly across different datasets due to varying data lengths. For these two metrics, we provide an approximate mean value of each model across all datasets. `-' indicates not applicable or not involved.
}
\begin{tabular}{
p{1.5cm}< \centering| 
p{2.9cm}< \centering| p{2.3cm}< \centering p{1.6cm}< \centering
p{1.4cm}< \centering p{1.4cm}< \centering  p{1.4cm}< \centering 
p{0.6cm}< \centering p{0.6cm}< \centering}
\toprule[1pt]
\textbf{Modality} &
\textbf{Models} & \textbf{PLLMs} & \textbf{PVM} &
\textbf{To-Paras} & \textbf{Tr-Paras} & \textbf{Held-In}
& \textbf{GPU} & \textbf{Time}\\
\midrule[1pt]
\multirow{6}{*}{\textbf{Text}}
& \textbf{ChatGPT} & gpt-3.5-turb &	- & - & - & -& - & -\\
& \textbf{LLaMA1-7B}& LLaMA-V1-7B & - & 6.74B & 6.74B & - & 26G & 2.0s \\
& \textbf{LLaMA1-13B} & LLaMA-V1-13B & - & 13.02B & 13.02B & - & 48G & 9.0s\\
& \textbf{LLaMA2-7B} & LLaMA-V2-7B & - & 6.74B & 6.74B  & - &26G & 1.0s \\
& \textbf{LLaMA2-13B} & LLaMA-V2-13B & - & 13.02B & 13.02B& - & 48G & 8.0s\\
& \textbf{Flan-T5-XXL}	& FlanT5-XXL & - & 11.14B & 11.14B & - & 44G & 0.3s  \\
\midrule[1pt]
\multirow{13}{*}{\textbf{Multimodal}}
& \textbf{OpenFlamingo} &  LLaMA-7B & ViT-L/14 & 8.34B & 1.31B & & 34G & 1.5s\\
& \textbf{Fromage} & OPT-6.7B & ViT-L/14 & 6.97B & 0.21B & - &14G  & 5.0s\\
& \textbf{LLaVA-7B} & LLaMA-7B & ViT-L/14 & 6.74B	& 6.74B	& ScienceQA & 15G & 2.5s\\
& \textbf{LLaVA-13B} & LLaMA-13B & ViT-L/14 & 13.02B & 13.02B & ScienceQA & 27G & 2.0s\\ 
& \textbf{MiniGPT4} & Vicuna-13B &  ViT-g/14 & 14.11B & 0.04B & - & 15G & 1.3s\\
& \textbf{mPLUG-Owl} & LLaMA-7B & ViT-L/14 & 7.12B & 7.12B	& - &16G & 4.0s \\
& \textbf{LLaMA-Adapter V2} & LLaMA-7B & ViT-L/14 & 7.23B & 7.23B & - & 14G & 1.3s\\
& \textbf{VPGTrans} & Vicuna-7B & -  & 7.83B &	0.11B & - & 36G & 10s\\
& \textbf{Multimodal-GPT} &  LLaMA-7B & ViT-L-14 & 8.37B & 0.02B & - &	18G	& 0.5s \\
& \textbf{LaVIN-7B} & LLaMA-7B & ViT-L/14 & 7.17B &	7.17B & ScienceQA & 16G & 4.0s  \\
& \textbf{LaVIN-13B} &  LLaMA-13B & ViT-L/14 & 13.36B & 13.36B	& ScienceQA & 28G& 11.0s \\
& \textbf{Lynx} & Vicuna-7B & Eva-ViT-1b & 8.41B & 0.69B & Hate & 44G & 6.5s \\
\midrule
\multirow{2}{*}{\textbf{Multimodal}}
& \textbf{BLIP-2} & FlanT5-XXL & ViT-g/14 & 12.23B & 0.11B	&  - & 26G & 3.5s\\
& \textbf{InstructBLIP} & FlanT5-XXL & ViT-g/14 & 12.31B & 0.45B & - & 16G	& 0.3s\\
\bottomrule[1pt]
\end{tabular}
\label{Table_different_models}
\end{center}
\end{table*}

\subsection{Multimodal Aspect-Based Sentiment Analysis} 
Multimodal Aspect-Based Sentiment Analysis (MABSA) devotes to detecting sentiments for specific aspect terms, dependent on the corresponding text-image context \cite{DBLP:conf/acl/HuPHLL19, DBLP:conf/ijcai/Yu019, DBLP:conf/emnlp/JuZXLLZZ21, DBLP:conf/acl/LingYX22, DBLP:conf/emnlp/YangZ022, DBLP:journals/taffco/YuCX23, DBLP:journals/ipm/YangNY22, DBLP:conf/mm/YuZL22}. We conduct experiments using three widely recognized datasets, namely, \textbf{Twitter-2015} \cite{DBLP:conf/aaai/0001FLH18}, \textbf{Twitter-2017} \cite{DBLP:conf/acl/JiZCLN18}, and \textbf{MASAD} \cite{DBLP:journals/ijon/ZhouZHHH21}.
\subsection{Multimodal Hateful Memes Recognition}
A new challenge set for multimodal classification is introduced, specifically designed to address Multimodal Hateful Memes Recognition (MHMR) in \citet{mathias2021findings}. However, no labeled test set has been publicly released for this challenge. Therefore, we assess on this task using the publicly labeled validation set, commonly referred to as`dev-seen' in the literature. For convenience, we refer to this dataset as \textbf{Hate} in this paper.
\subsection{Multimodal Sarcasm Recognition}
Multimodal Sarcasm Recognition (MSR) focuses on identifying sarcasm in multimodal content \cite{DBLP:conf/acl/CaiCW19, DBLP:conf/acl/LiangLLY00PX22, DBLP:conf/aaai/QiaoJSCZN23}. 
\citet{DBLP:conf/acl/CaiCW19} introduce a new dataset specifically for multimodal Twitter sarcasm detection, known as \textbf{Sarcasm} in our paper. 
\subsection{Multimodal Relation Extraction}
The Multimodal Relation Extraction (MRE) task entails the identification of textual relations between two entities with the assistance of visual content \cite{DBLP:conf/mm/ZhengFFCL021, DBLP:conf/coling/0023HDWSSX22, DBLP:conf/aaai/Yuan0WL23}. To facilitate research in this domain, \citet{DBLP:conf/mm/ZhengFFCL021} introduce the multimodal neural relation extraction dataset (\textbf{MNRE}), which is manually labeled and serves as a valuable resource for the MRE task.
\subsection{Visual Question Answering}
\textbf{ScienceQA} \cite{DBLP:conf/nips/LuMX0CZTCK22} is a popular visual question-answering dataset with diverse science topics that provide both image context and text context, which can be either semantically rich information or a simple hint. Our focus is on multimodal evaluation, so we only utilize the portion that includes image context.
\section{Model Details}
\label{Appendix_large_models}

\begin{figure*}[t] 
  \centering 
  \includegraphics[scale = 0.6]{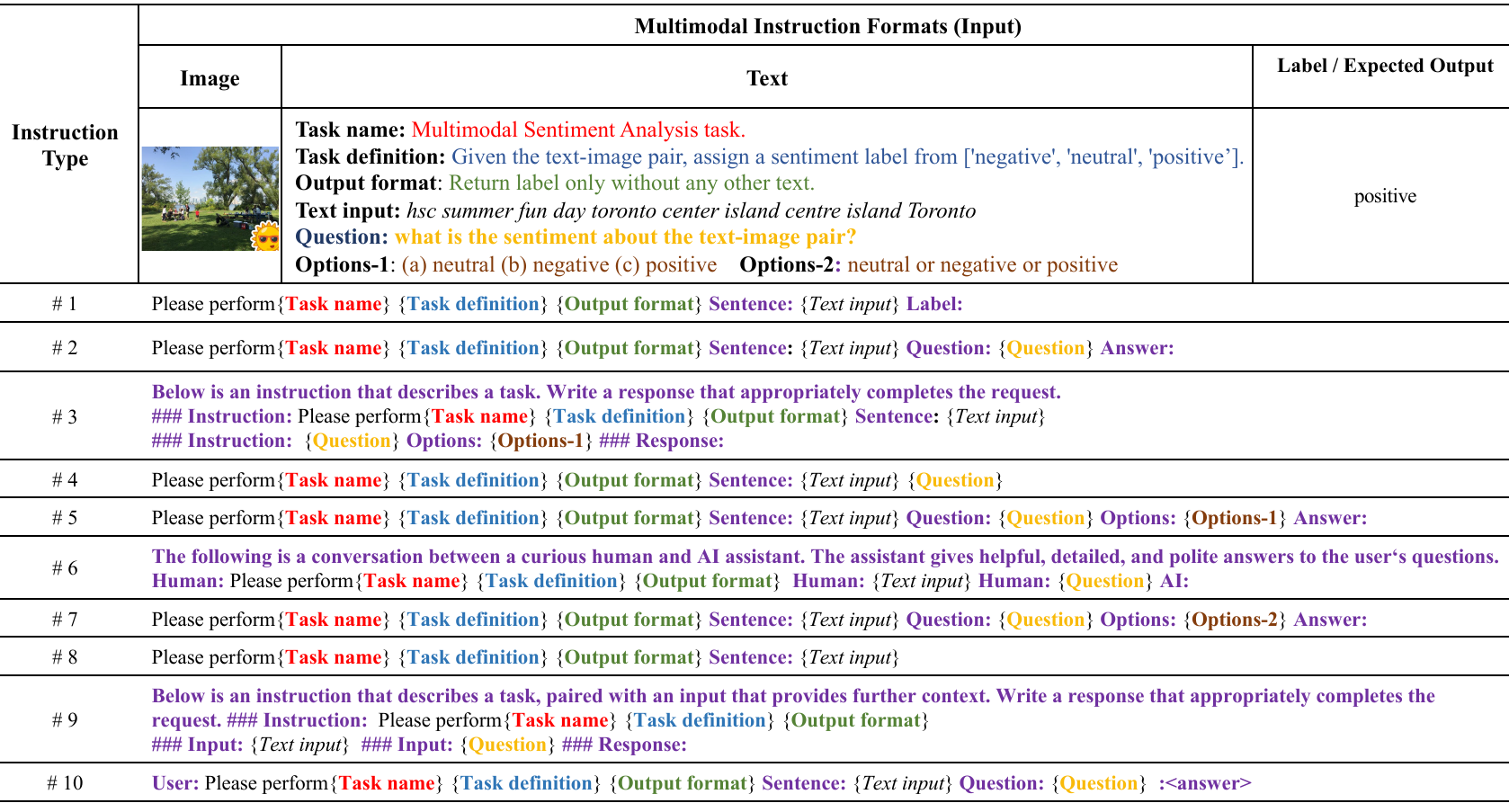} 
  \caption{The column labeled "Multimodal Instruction Formats" displays the formats of different instructions designed specifically for the MSA task, taking the MVSA-Single dataset as an example. Each instruction includes the text context, such as \textit{Text input}; Image context; essential components, including \textbf{\textcolor{red}{Task name}}, \textbf{\textcolor[RGB]{70,130,180}{Task definition}}, \textbf{\textcolor[RGB]{34,139,34}{Output format}}, and \textbf{\textcolor[RGB]{153,50,204}{Specific instruction trigger words}}; and optional components, such as \textbf{\textcolor[RGB]{255,215,0}{Question}} and various \textbf{\textcolor[RGB]{205,133,63}{Options}}.
   }
  \label{Figure_1_MSA_Multimodal_Instruction} 
\end{figure*}
We assess a diverse set of Language Models (LMs), including \textbf{6} pure Large Language Models (LLMs) and  \textbf{14} Multimodal Large Language Models (MLLMs). Details of the different models are provided in Table \ref{Table_different_models}.
\subsection{LLMs} 
\label{Appendix_LLMs}
We assess various LLMs across multiple tasks using text-only content.
\textbf{ChatGPT} \cite{chatgpt} is a conversational AI language model developed by OpenAI,  known for its impressive performance across a wide range of NLP tasks. Specifically, we assess the classic model, ``ChatGPT (gpt-3.5-turbo)"\footnote{Our evaluation of ChatGPT is carried out between July and September 2023. Currently, the GPT-4 API only accepts plain text requests (image inputs are still in limited alpha), and it uses the same ChatCompletions API as gpt-3.5-turbo, as stated on the \url{https://openai.com/research/gpt-4}. Consequently, we do not assess GPT4 in our evaluation.}, using the official API.
\textbf{Flan-T5} \cite{DBLP:journals/corr/abs-2210-11416} is a model that extends its capabilities to 1836 fine-tuning tasks through instruction tuning, enhancing the performance and usability of the model. The specific version we examine in this evaluation is ``flan-t5-xxl".
The LLaMA family of models includes \textbf{LLaMA-1} \cite{DBLP:journals/corr/abs-2302-13971} and \textbf{LLaMA-2} \cite{DBLP:journals/corr/abs-2307-09288} models.
We evaluate the performance of the ``decapoda-llama-7b/13b-hf" and ``meta-Llama-2-7b/13b-hf" models.
\begin{figure*}[t] 
  \centering 
 
  \includegraphics[scale = 0.65]{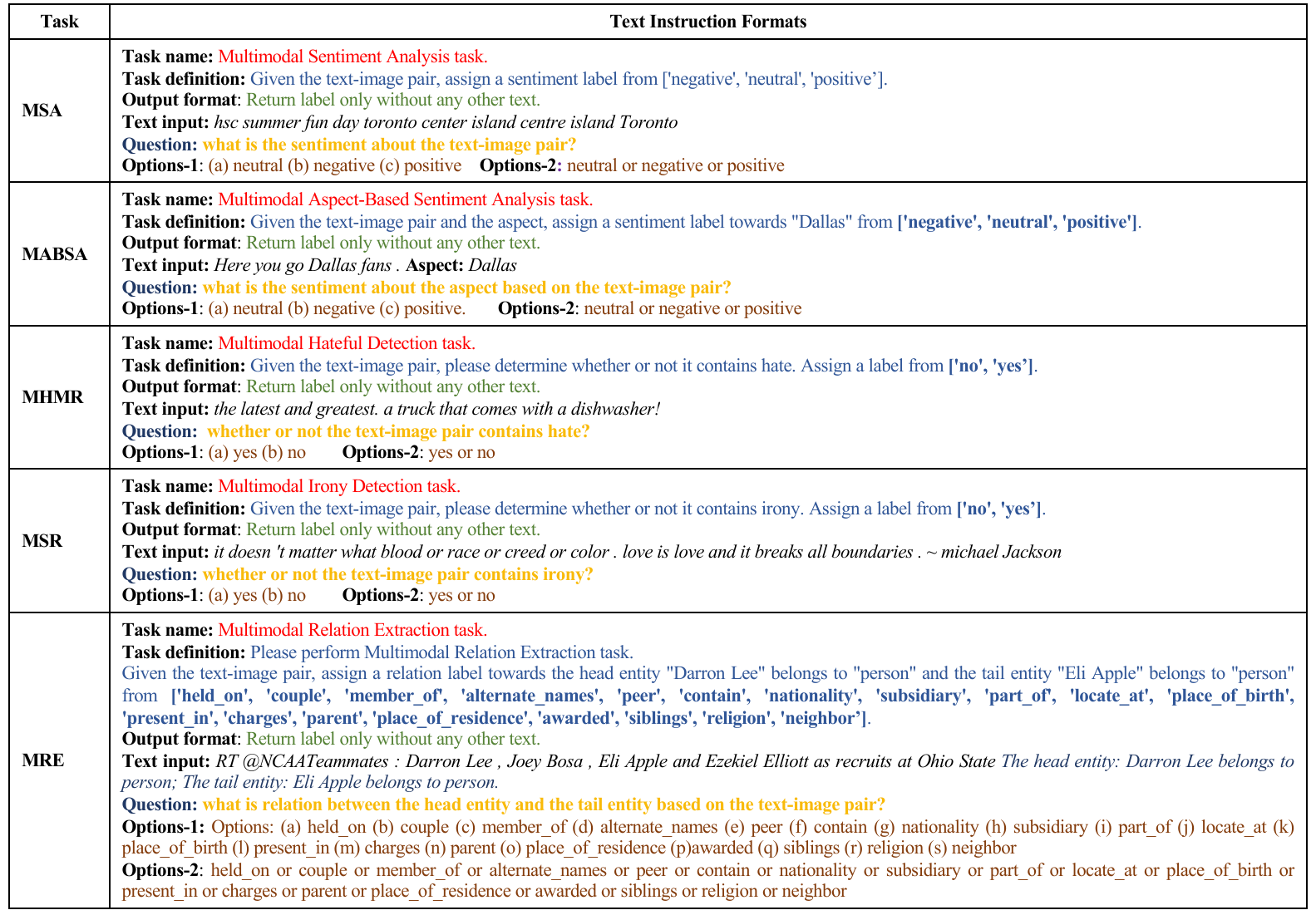} 
   \caption{Details of multiple instruction components for different tasks, such as Multimodal Sentiment Analysis (MSA), Multimodal Aspect-Based Sentiment Analysis (MABSA), Multimodal Hateful Memes Recognition (MHMR), Multimodal Sarcasm Recognition (MSR), and Multimodal Relation Extraction (MRE).
   }
  \label{Figure_2_text_instruction_for_different_tasks} 
\end{figure*}

\subsection{MLLMs}
\label{Appendix_MLLMs}
The MLLM is trained by combining a pretrained visual model (PVM) and a pre-trained LLM. Commonly, the PVM is derived from pre-trained CLIP models, such as ViT-L/14, ViT-g/14, and etc. MLLMs can be categorized into the LLaMA family, including LLaMA-V1, LLaMA-V2, and Vicuna, as well as the FlanT5 family.  It's important to note that, for the sake of fairness in our evaluation, all LLaMA models used in the MLLMs are from the LLaMA-1 series.
We provide a comprehensive overview of MLLMs that are based on the LLaMA-1 architecture as the LLM.
We evaluate the pre-trained model ``Openflamingo-9B", which is part of the \textbf{OpenFlamingo} family. This family of models demonstrate impressive performance in few-shot learning across various open-ended vision and language tasks and is referenced in \cite{DBLP:conf/nips/AlayracDLMBHLMM22, anas_awadalla_2023_7733589}.
\textbf{Fromage} is trained by visually grounding LLMs through image captioning and contrastive learning \cite{koh2023grounding}. We assess the pre-trained  ``fromage-model''.
We assess different pretrained LLaVA models, which are based on varying scale Large Language Models (LLMs), including ``LLaVA-7B" and ``LLaVA-13B." \textbf{LLaVA} \cite{DBLP:journals/corr/abs-2304-08485}, is an end-to-end trained large multimodal model that connects a vision encoder with an LLM for comprehensive visual and language understanding.
\citet{DBLP:journals/corr/abs-2304-10592} introduce \textbf{MiniGPT-4} aligns a frozen visual encoder with a frozen LLM called Vicuna, using just one projection layer. We utilize the pretrained ``MiniGPT-4 checkpoint with Vicuna 13B''.
\textbf{mPLUG-Owl} equips Large Language Models (LLMs) with multi-modal abilities through modularized learning of a foundation LLM, a visual knowledge module, and a visual abstractor module \cite{DBLP:journals/corr/abs-2304-14178}. We evaluate the ``mPLUG-Owl 7B'' model.
\textbf{LLaMA-Adapter V2} \cite{DBLP:journals/corr/abs-2304-15010} are jointly trained on image-text pairs and instruction-following data. We evaluate the ``LLaMA-Adapter V2 Multimodal'' model.
\citet{DBLP:journals/corr/abs-2305-01278} present a two-stage transfer framework, \textbf{VPGTrans}, designed to achieve efficient visual prompt generator (VPG) transfer across LLMs while using less training data. We assess the performance of the ``VL-LLaMA'' model, where the VPG is transferred from BLIP-2 OPT6.7B to LLaMA-7B.
\textbf{Multimodal-GPT} \cite{DBLP:journals/corr/abs-2305-04790} incorporates the Low-rank Adapter (LoRA\cite{DBLP:conf/iclr/HuSWALWWC22}) in both the gated-cross-attention and self-attention components of the language model. We asses the pretrained ``mmgpt-lora-v0-release" weights.
\textbf{LaVIN} \cite{DBLP:journals/corr/abs-2305-15023} is a model proposed based on the concept of Mixture-of-Modality Adaptation (MMA). We conduct evaluation of the ``LaVIN-7B'' and ``LaVIN-13B''.
\textbf{Lynx} \cite{DBLP:journals/corr/abs-2307-02469} is a straightforward prefix-tuning GPT4-style model featuring a two-stage training approach. We utilize the ``finetune-lynx.pt" model for evaluation.
We further introduce MLLMs based on the Flan-t5 text backbone. 
\textbf{BLIP-2} \cite{DBLP:journals/corr/abs-2301-12597} addresses the modality gap through the use of a lightweight Querying Transformer that connects a frozen pre-trained image model with a language model. We consider the version of this model labeled as ``blip2-flan-t5-xxl".
\textbf{InstructBLIP} \cite{DBLP:journals/corr/abs-2305-06500} conducts a comprehensive and systematic study on vision-language instruction tuning by utilizing pretrained BLIP-2 models.  We utilize the pretrained model ``blip2-instruct-flant5xxl".

\section{Multimodal Instructions}
\label{multimodal instructions}
\subsection{Multimodal Instruction for the Multimodal Sentiment Analysis Task}
\label{Appendix_MSA_Multimodal_Instructions}
We design different multimodal instructions for the multimodal sentiment analysis task and take the `MVSA-Single' dataset as an example, as illustrated in Figure \ref{Figure_1_MSA_Multimodal_Instruction}.
Each instruction consists of the text context, such as \textit{Text input}; Image context; essential components, including \textbf{\textcolor{red}{Task name}}, \textbf{\textcolor[RGB]{70,130,180}{Task definition}}, \textbf{\textcolor[RGB]{34,139,34}{Output format}}, and \textbf{\textcolor[RGB]{153,50,204}{Specific instruction trigger words}}; and optional components, such as \textbf{\textcolor[RGB]{255,215,0}{Question}} and various \textbf{\textcolor[RGB]{205,133,63}{Options}}.
Except for ScienceQA, other tasks do not include the ``Context'' component; instead, they are replaced by ``Text input''as the text context for each instance, as the Figure \ref{Figure_1_MSA_Multimodal_Instruction} shows. Furthermore, the ``Question'' component is optional and is used to simulate the visual question answering task.
\subsection{Different Instructions for Various Multimodal Content Comprehension Tasks}
\label{Appendix_different_instructions_for_multiple_taks}
Similar to MSA, designed multimodal instructions for other tasks also include the text context, such as \textit{Text input}; Image context; essential components, including \textbf{\textcolor{red}{Task name}}, \textbf{\textcolor[RGB]{70,130,180}{Task definition}}, \textbf{\textcolor[RGB]{34,139,34}{Output format}}, and \textbf{\textcolor[RGB]{153,50,204}{Specific instruction trigger words}}; and optional components, such as \textbf{\textcolor[RGB]{255,215,0}{Question}} and various \textbf{\textcolor[RGB]{205,133,63}{Options}}, as shown in the Figure \ref{Figure_2_text_instruction_for_different_tasks}.

\section{Comparison of Experimental Results in ScienceQA (with/without Text Context)}
\label{Appendix_Comparison_of_ScienceQA}
ScienceQA primarily answers questions based on the image modality. Therefore, we also conduct comparative experiments without text context, where there is no `Context' item in Figure \ref{Figure_1_ScienceQA_different_instructions}. The experimental results are displayed in Table \ref{Table_ScienceQA}. We find that most MLLMs performs better on ScienceQA with text context; the image modailty is more important for ScienceQA, such as the performance of Flan-T5-XXL is 6.7\% lower than BLIP2 and 5.9\% lower than InstructBLIP; when text context is removed, their performance decreases by 4.46\% and 5.01\%, respectively.

\begin{table*}[!htp]\small
\begin{center}
\renewcommand{\arraystretch}{1} 
\caption{
Comparison of experimental results in ScienceQA (with/without Text Context) for best performance , $A^{\tilde{i}}$ ($\uparrow$, measured by Accuracy), and Mean Relative Gain of MLLMs,  MRG$^{\mathcal{M}}$ ($\uparrow$), across all instructions. $\nabla$ represents the difference in experimental results between with text context and without text context in ScienceQA. `w/o' means without.
}
\begin{tabular}{
p{1.8cm}< \centering| 
p{1.2cm}< \centering p{4cm}< \centering p{1.2cm}< \centering| 
p{1.2cm}< \centering p{4cm}< \centering
}
\toprule[1pt]
\multirow{2}{*}{\textbf{Models}}
& \multicolumn{3}{c|}{\textbf{Best Performance}}  
& \multicolumn{2}{c}{\textbf{Mean Relative Gain}} \\
\cline{2-6}
& \textbf{ScienceQA} & \textbf{ScienceQA (w/o Text Context)} & $\nabla$ & \textbf{ScienceQA} & \textbf{ScienceQA (w/o Text Context)}\\
\midrule[1pt]
\textbf{OpenFlamingo} & 39.27$^5$ & 41.05$^5$ & -1.78 &-54.51 & -50.41 \\
\textbf{Fromage} & 34.51$^7$ & 35.90$^7$ & -1.39  & -66.76 & -58.03\\
\textbf{LLaVA-7B} & 41.10$^5$ & 43.03$^5$ & -1.93 &  1.76   & 5.45\\
\textbf{LLaVA-13B} & 47.74$^7$ & 45.61$^7$ & 2.13  &  8.58  &  3.37 \\
\textbf{MiniGPT4} & 58.70$^5$ & 58.70$^5$ & 0 & 27.56 & 25.00 \\
\textbf{mPLUG-Owl} & 37.93$^7$ & 36.64$^6$ & 1.29 & -31.34 & -41.56\\
\textbf{AdapterV2}  & 54.44$^7$ & 53.64$^7$ & 0.80 & 35.82$^\star$ & 33.64$^\star$\\
\textbf{VPGTrans}  & 47.00$^3$ & 45.71$^3$ & 1.29 & -20.49 & -13.54\\
\textbf{MultiGPT} & 36.29$^5$ & 38.82$^5$ & -2.53 & -56.34 & -45.61\\
\textbf{LaVIN-7B}  & 75.11$^3$ & 71.69$^3$ & 3.42  & 1.50 &   -1.43\\
\textbf{LaVIN-13B}  &  \textbf{77.54$^5$} & \textbf{73.47$^5$}  & 4.07 & 7.00 &  6.93\\
\textbf{Lynx} & 38.28$^7$ & 43.53$^6$ & -5.25 & -46.08 & -31.96\\
\midrule[1pt]
\textbf{BLIP-2}  & 74.17$^1$ & 69.71$^2$ & 4.46 & \underline{93.34} & \textbf{85.74}\\
\textbf{InstructBLIP} & 73.33$^2$ & 68.32$^2$ & 5.01 & \textbf{99.94} & \underline{82.39}\\
\bottomrule[1pt]
\end{tabular}

\label{Table_ScienceQA}
\end{center}
\end{table*}

\begin{table*}[!htp]\small
\begin{center}
\renewcommand{\arraystretch}{1} 
\caption{
The best zero-shot performance, $A^{\tilde{i}}$ ($\uparrow$, measured by Accuracy), of various LMs on MOSI-2, MOSI-7, MOSEI-2, and MOSEI-7 datasets, with superscripts indicating the specific instruction 
that resulted in the best performance for each dataset on the respective model. The ``\textbf{Total}" column represents the sum of accuracy scores across four datasets for each model.
}
\begin{tabular}{
p{1.8cm}< \centering| 
p{1.2cm}< \centering p{1.2cm}< \centering p{1.2cm}< \centering p{1.2cm}< \centering|
p{1cm}< \centering
}
\toprule[1pt]
\multirow{2}{*}{\textbf{Models}}
& \multicolumn{4}{c|}{\textbf{MSA}}  
& \multirow{2}{*}{\textbf{Total}} \\
\cline{2-5}
& \textbf{MOSI-2} & \textbf{MOSI-7} & \textbf{MOSEI-2} & \textbf{MOSEI-7} \\
\midrule[1pt]
\textbf{ChatGPT} & \textbf{89.60$^5$} & \textbf{44.44$^{10}$} & 84.97$^5$ & 40.77$^1$ & 259.78\\
\textbf{LLaMA1-7B}  & 82.01$^2$ & 34.26$^2$ & 75.62$^1$ &  15.50$^1$ & 207.39 \\
\textbf{LLaMA1-13B}  & 72.10$^5$ & 34.11$^2$ & 79.55$^2$ & 28.74$^2$ & 214.50 \\
\textbf{LLaMA2-7B}  & 67.68$^1$ & 26.38$^1$ & 77.30$^1$ & 16.78$^1$ & 188.14 \\
\textbf{LLaMA2-13B}  & 81.86$^2$ & 31.49$^6$ & 81.66$^2$ & 24.33$^6$ & 219.34 \\
\textbf{Flan-T5-XXL}  &  \textbf{89.60$^{10}$}  & 42.86$^6$ & 86.52$^6$ & \textbf{46.29$^6$} & \textbf{265.27}\\
\midrule[1pt]
\textbf{OpenFlamingo} & 79.97$^7$ & 24.85$^2$ & 77.3$^7$ & 12.12$^2$ &  194.24  \\
\textbf{Fromage} & 57.19$^7$ & 19.15$^2$ & 47.41$^2$ & 11.04$^2$ & 134.79 \\
\textbf{LLaVA-7B} & 74.69$^2$ & 30.03$^9$ & 74.65$^7$ & 18.12$^9$ & 197.49  \\
\textbf{LLaVA-13B} &  80.18$^7$ & 30.90$^6$ &  76.58$^7$ & 28.37$^3$ & 216.03 \\
\textbf{MiniGPT4} & 83.99$^4$ & 35.42$^2$ & 83.38$^2$ &  38.46$^5$   &  241.25 \\
\textbf{mPLUG-Owl} & 68.75$^1$ & 28.28$^6$ & 58.10$^6$ &  20.29$^6$ & 175.42 \\
\textbf{AdapterV2}  & 86.43$^8$ & 38.34$^8$ & 82.02$^8$ & 33.53$^9$ & 240.32\\
\textbf{VPGTrans}  & 76.22$^4$ & 30.47$^4$ & 76.76$^4$ & 38.27$^6$ &  221.72\\
\textbf{MultiGPT} & 68.35$^7$ & 25.58$^2$ & 72.76$^7$ & 10.17$^5$ &  176.86\\
\textbf{LaVIN-7B}  & 71.41$^5$ & 25.73$^5$ & 69.97$^7$ & 29.46$^1$ &  196.57\\
\textbf{LaVIN-13B}  & 79.97$^7$ & 27.63$^1$ & 73.54$^7$ & 27.20$^7$ &  208.34  \\
\textbf{Lynx}  & 74.77$^7$ & 22.37$^2$ & 73.72$^7$ & 10.28$^2$ & 181.14 \\
\midrule[1pt]
\textbf{BLIP-2}  & 88.99$^9$ & 43.42$^2$ & \textbf{86.88$^6$} & 45.79$^6$ & 265.08\\
\textbf{InstructBLIP}  & 88.68$^9$ & 43.28$^2$ & 85.98$^9$ & 45.68$^9$ &  263.62\\
\bottomrule[1pt]
\end{tabular}

\label{Table_Best_Acc_MOSI_MOSEI}
\end{center}
\end{table*}

\begin{table*}[!htp]\small
\begin{center}
\renewcommand{\arraystretch}{1} 
\caption{
The mean relative gain, MRG$^{\mathcal{M}}$ ($\uparrow$), of MOSI-2, MOSI-7, MOSEI-2, and MOSEI-7 datasets for various instructions across all models.
}
\begin{tabular}{
p{1.7cm}< \centering| 
p{1.2cm}< \centering p{1.2cm}< \centering p{1.2cm}< \centering p{1.2cm}< \centering|
p{1cm}< \centering p{1cm}< \centering
}
\toprule[1pt]
\multirow{2}{*}{\textbf{Models}}
& \multicolumn{4}{c|}{\textbf{MSA}}  
& \multirow{2}{*}{\textbf{Wins1}} & \multirow{2}{*}{\textbf{Wins3}} \\
\cline{2-5}
& \textbf{MOSI-2} & \textbf{MOSI-7} & \textbf{MOSEI-2} & \textbf{MOSEI-7} \\
\midrule[1pt]
\textbf{ChatGPT} & 56.47$^\star$ &   \textbf{96.16} &   47.59 &  107.61 & \textbf{1} & \textbf{2} \\
\textbf{LLaMA1-7B}  & -16.45 &  -12.51 &  -21.11 &  -44.64 & 0 & 0 \\
\textbf{LLaMA1-13B}  &  -12.74 &   -8.46 &   -6.36 &  -32.84 & 0 & 0 \\
\textbf{LLaMA2-7B}  & -25.67 &  -33.67 &   -4.06 &  -50.26 & 0 & 0 \\
\textbf{LLaMA2-13B}  & -14.03 &  -33.19 &  -23.00  &    -55.47 & 0 & 0 \\
\textbf{Flan-T5-XXL}  & \underline{57.05}  &  85.86$^\star$ &   \underline{56.97} &  \textbf{151.73} & \textbf{1} & \textbf{4} \\
\midrule[1pt]
\textbf{OpenFlamingo} &  -30.84 &  -55.07 &  -37.34 &  -62.44 & 0 & 0 \\
\textbf{Fromage} &  -68.63 &  -82.11 &  -73.37 &  -88.11 & 0 & 0 \\
\textbf{LLaVA-7B} &  8.91 &   10.25 &   10.55 &  -16.04 & 0 & 0  \\
\textbf{LLaVA-13B} &   8.72 &   -6.01  &  14.06  &   5.95 &  0 & 0 \\
\textbf{MiniGPT4} &  29.79  &  38.47  &  36.37  &  35.35 & 0 & 0  \\
\textbf{mPLUG-Owl} &  -23.83  & -23.37  & -28.72  & -41.64 & 0 & 0 \\
\textbf{AdapterV2}  & 9.27  &  15.76  &  23.19  & -11.11 & 0 & 0 \\
\textbf{VPGTrans}  &   -7.97  & -13.85  &  -4.34   & -5.15 & 0 & 0  \\
\textbf{MultiGPT} &  -43.19  & -43.79  & -47.15  & -70.15 & 0 & 0\\
\textbf{LaVIN-7B}  &  -22.11  & -31.48  & -37.1   & -44.85 & 0 & 0\\
\textbf{LaVIN-13B}  & 1.10  &  -12.14   & -8.20  & -3.26 & 0 & 0 \\
\textbf{Lynx}  &  -19.11 &  -58.18 &  -12.93 &  -67.22 & 0 & 0 \\
\midrule[1pt]
\textbf{BLIP-2}  &  \textbf{57.17} &   \underline{87.89} &   \textbf{58.49} &  \underline{148.67}  & \textbf{2} & \textbf{4}\\
\textbf{InstructBLIP}  & 56.10 &    79.42 &   56.47$^\star$ &  143.86$^\star$ & 0 & \textbf{2} \\
\bottomrule[1pt]
\end{tabular}
\label{Table_Instructions_Mean_Relative_Gain_MOSI_MOSEI}
\end{center}
\end{table*}

\section{Experimental Results on MOSI and MOSEI datasets}
\label{Appendix_Results_MOSI_MOSEI}
\subsection{The Best Performance}
\label{Appendix_Best_Acc_MOSI_MOSEI}
The best performance of MOSI and MOSEI datasets are shown in Table \ref{Table_Best_Acc_MOSI_MOSEI}.
Different from other datasets, ChatGPT performs best on MOSI-2 and MOSI-7 datasets. Flan-T5-XXL, BLIP2, and InstructBLIP have the similar performance on four datasets. It could be attributed to the fact that the videos in the MOSI and MOSEI datasets consist of self-portraits of individuals, and randomly selected image frames may not effectively capture the key frames that convey the emotions in the video, consequently not yielding any advantage for MLLMs. The evaluation of video datasets using MLLMs remains a significant challenge and represents one of our future research directions.
\subsection{The Mean Relative Gain for Various LMs}
\label{Appendix_Mean_Relative_Gain_for_Various LMs}
The mean relative gain of MOSI and MOSEI datasets for various LMs are shown in Table \ref{Table_Model_Mean_Relative_Gain_MOSI_MOSEI}.
Although ChatGPT can achieve the best performance with certain instructions, its mean relative gain performance is not as good as Flan-T5-XXL and BLIP2. This indicates that ChatGPT excels with specific instructions but struggles with others, showing higher sensitivity to instruction variations. On the other hand, Flan-T5-XXL and BLIP2 demonstrate greater stability across different instructions.

\begin{table*}[!htp]\small
\begin{center}
\renewcommand{\arraystretch}{1} 
\caption{
The mean relative gain, MRG$^{\mathcal{I}}$ ($\uparrow$), of MOSI-2, MOSI-7, MOSEI-2, and MOSEI-7 datasets for various models across instructions.
}
\begin{tabular}{
p{1.7cm}< \centering| 
p{1.2cm}< \centering p{1.2cm}< \centering p{1.2cm}< \centering p{1.2cm}< \centering|
p{1cm}< \centering p{1cm}< \centering
}
\toprule[1pt]

\multirow{2}{*}{\textbf{Instructions}}
& \multicolumn{4}{c|}{\textbf{MSA}}  
& \multirow{2}{*}{\textbf{Wins1}} & \multirow{2}{*}{\textbf{Wins3}} \\
\cline{2-5}
& \textbf{MOSI-2} & \textbf{MOSI-7} & \textbf{MOSEI-2} & \textbf{MOSEI-7} \\
\midrule[1pt]
\textbf{\# 1} & \underline{30.94} & \underline{36.24} &  20.94$^\star$ &  \underline{32.88} & 0 & \textbf{4} \\
\textbf{\# 2} & \textbf{31.02} &  \textbf{56.91} &  \underline{29.91} &  \textbf{49.86} & \textbf{3} & \textbf{4} \\
\textbf{\# 3} & -2.96 & -16.44 &  -7.04 &  -4.66 & 0 & 0 \\
\textbf{\# 4}  & -20.55 & -10.39 & -16.32 & -11.89 & 0 & 0 \\
\textbf{\# 5}  & 12.39  &  1.47 &   5.09 &  -0.02 & 0 & 0 \\
\textbf{\# 6}  & -16.55 &  -1.16 &  -2.60 &  16.57$^\star$ & 0 & \textbf{1} \\
\textbf{\# 7}  & 26.55$^\star$ & -27.73 &  \textbf{37.76} & -24.32 & \textbf{1} & \textbf{2} \\
\textbf{\# 8} & -29.25 & -17.70  &  -28.22 & -27.59 & 0 & 0 \\
\textbf{\# 9} &  5.26 &  16.29$^\star$ &  -2.40  & 11.38 & 0 & \textbf{1} \\
\textbf{\# 10} & -36.85 & -37.49 & -37.14 & -42.21 & 0 & 0  \\
\bottomrule[1pt]
\end{tabular}

\label{Table_Model_Mean_Relative_Gain_MOSI_MOSEI}
\end{center}
\end{table*}
\subsection{The Mean Relative Gain for Various Instructions}
\label{Appendix_Mean_Relative_Gain_for_Various_Instructions}
The mean relative gain of MOSI and MOSEI datasets for various instructions are shown in Table \ref{Table_Instructions_Mean_Relative_Gain_MOSI_MOSEI}.
Instruction \# 2, in the Question-Answering format with the `options' term, delivers the best performance on the MOSI and MOSEI datasets. Particularly for the more finely classified MOSI-7 and MOSEI-7 datasets, the `options' term may introduce more complexity to the model and potentially diminish its performance.
\subsection{The Stability of Models}
\label{Appendix_the_stability_LMs}
The stability of MOSI and MOSEI datasets for excellent models are shown in Table \ref{Table_stability_of_various_large_models_on_MOSI_MOSEI}.
The performance of models on fine-grained classification datasets, like MOSI-7 and MOSEI-7, exhibits greater fluctuations compared to coarse-grained datasets, such as MOSI-2 and MOSEI-2.

\begin{table*}[h] \small
\begin{center}
\renewcommand{\arraystretch}{1} 
\caption{The stability, $S^{\mathcal{M^{'}}}$ ($\downarrow$), of various models with excellent performance across instructions on MOSI-2, MOSI-7, MOSEI-2, and MOSEI-7 datasets. 
}
\begin{tabular}{
p{1.8cm}< \centering| 
p{1.2cm}< \centering p{1.2cm}< \centering p{1.2cm}< \centering p{1.2cm}< \centering|
p{1cm}< \centering
}
\toprule[1pt]
\textbf{Models} 
& \textbf{MOSI-2} & \textbf{MOSI-7} & \textbf{MOSEI-2} & \textbf{MOSEI-7} & 
\textbf{Wins1} \\
\midrule[1pt]
\textbf{ChatGPT} & 1.56 &  \textbf{2.17} & 5.9 &  3.57 & \textbf{1} \\
\textbf{Flan-T5-XXL} & 0.73  & 3.73 & \textbf{0.38} & \textbf{2.97} & \textbf{2}\\
\midrule[1pt]
\textbf{BLIP-2} & \textbf{0.31} & 3.62  & 0.42  & 3.43  & \textbf{1} \\
\textbf{InstructBLIP} & 0.62 &  5.02 &  0.51 &  4.24 & 0 \\
\bottomrule[1pt]
\end{tabular}

\label{Table_stability_of_various_large_models_on_MOSI_MOSEI}
\end{center}
\end{table*}

\subsection{The Stability of Instructions}
\label{Appendix_the_stability_instructions}
The stability of MOSI and MOSEI datasets for different instructions are shown in Table \ref{Table_stability_of various_instructions_on_MOSI_MOSEI}. Similar to other datasets, Instruction \# 2 performs better on the MOSI and MOSEI datasets, particularly on fine-grained datasets.

\begin{table*}[!htp] \small
\begin{center}
\renewcommand{\arraystretch}{1} 
\caption{The stability, $S^{\mathcal{I^{'}}}$ ($\downarrow$), of different instructions with excellent performance across  models on MOSI-2, MOSI-7, MOSEI-2, and MOSEI-7 datasets.
}
\begin{tabular}{
p{1.8cm}< \centering| 
p{1.2cm}< \centering p{1.2cm}< \centering p{1.2cm}< \centering p{1.2cm}< \centering|
p{1cm}< \centering
}
\toprule[1pt]
\textbf{Instructions} 
& \textbf{MOSI-2} & \textbf{MOSI-7} & \textbf{MOSEI-2} & \textbf{MOSEI-7}
& \textbf{Wins1} \\
\midrule[1pt]
\textbf{\# 1} &  \textbf{11.71} & 10.05 & 16.36 & 13.25 & \textbf{1}\\
\textbf{\# 2} & 12.19 &  \textbf{8.89} & 15.77 & \textbf{11.93} & \textbf{2}\\
\textbf{\# 3} & 19.90 &  11.02 & 22.40  & 12.69 & 0\\
\textbf{\# 4} & 32.30 &  14.78 & 29.72 & 15.48 & 0\\
\textbf{\# 5} & 19.04 & 10.71 & 19.60 &  12.41 & 0\\
\textbf{\# 6} & 27.95 & 13.43 & 25.93 & 14.75 & 0\\
\textbf{\# 7} & 14.98 & 11.22 & \textbf{11.35} & 12.57 & \textbf{1} \\
\textbf{\# 9} & 22.24 & 11.98 & 24.45 & 14.25 & 0\\
\bottomrule[1pt]
\end{tabular}

\label{Table_stability_of various_instructions_on_MOSI_MOSEI}
\end{center}

\end{table*}

\end{document}